\documentclass[acmtog]{acmart}
\acmSubmissionID{580}

\usepackage{booktabs} 
\usepackage{lipsum}
\usepackage{soul,color} 
\usepackage{gensymb} 

\usepackage[norelsize, linesnumbered, ruled, lined, boxed, commentsnumbered, noend]{algorithm2e} 

\usepackage{caption}
\usepackage{subcaption}
\usepackage{multirow}
\usepackage[titletoc]{appendix}

\SetAlFnt{\small}
\SetAlCapFnt{\small}
\SetAlCapNameFnt{\small}
\SetAlCapHSkip{0pt}
\SetArgSty{textup}
\SetKwRepeat{Do}{do}{while}

\setcopyright{rightsretained}
\acmJournal{TOG}
\acmYear{2022}\acmVolume{41}\acmNumber{6}\acmArticle{278}\acmMonth{12} \acmDOI{10.1145/3550454.3555525}

\begin{document}
\title{Assemble Them All: Physics-Based Planning for Generalizable Assembly by Disassembly}

\author{Yunsheng Tian}
\orcid{0000-0002-6471-7575}
\affiliation{%
 \institution{MIT CSAIL}
 \country{USA}}
\email{yunsheng@csail.mit.edu}

\author{Jie Xu}
\orcid{0000-0003-3510-3387}
\affiliation{%
 \institution{MIT CSAIL}
 \country{USA}}
\email{jiex@csail.mit.edu}

\author{Yichen Li}
\orcid{0000-0002-5659-8748}
\affiliation{%
 \institution{MIT CSAIL}
 \country{USA}}
\email{yichenl@csail.mit.edu}

\author{Jieliang Luo}
\orcid{0000-0002-3298-9111}
\affiliation{%
 \institution{Autodesk Research}
 \country{USA}}
\email{rodger.luo@autodesk.com}

\author{Shinjiro Sueda}
\orcid{0000-0003-4656-498X}
\affiliation{%
 \institution{Texas A\&M University}
 \country{USA}}
\email{sueda@tamu.edu}

\author{Hui Li}
\orcid{0000-0001-7382-2425}
\affiliation{%
 \institution{Autodesk Research}
 \country{USA}}
\email{hui.xylo.li@autodesk.com}

\author{Karl D.D. Willis}
\orcid{0000-0002-6990-2294}
\affiliation{%
 \institution{Autodesk Research}
 \country{USA}}
\email{karl.willis@autodesk.com}

\author{Wojciech Matusik}
\orcid{0000-0003-0212-5643}
\affiliation{%
 \institution{MIT CSAIL}
 \country{USA}}
\email{wojciech@csail.mit.edu}

\begin{abstract}

Assembly planning is the core of automating product assembly, maintenance, and recycling for modern industrial manufacturing. Despite its importance and long history of research, planning for mechanical assemblies when given the final assembled state remains a challenging problem. This is due to the complexity of dealing with arbitrary 3D shapes and the highly constrained motion required for real-world assemblies. In this work, we propose a novel method to efficiently plan physically plausible assembly motion and sequences for real-world assemblies. Our method leverages the assembly-by-disassembly principle and physics-based simulation to efficiently explore a reduced search space. To evaluate the generality of our method, we define a large-scale dataset consisting of thousands of physically valid industrial assemblies with a variety of assembly motions required. Our experiments on this new benchmark demonstrate we achieve a state-of-the-art success rate and the highest computational efficiency compared to other baseline algorithms. Our method also generalizes to rotational assemblies (e.g., screws and puzzles) and solves 80-part assemblies within several minutes.

\end{abstract}

%
%


%
%


\begin{teaserfigure}
    \includegraphics[width=0.97\textwidth]{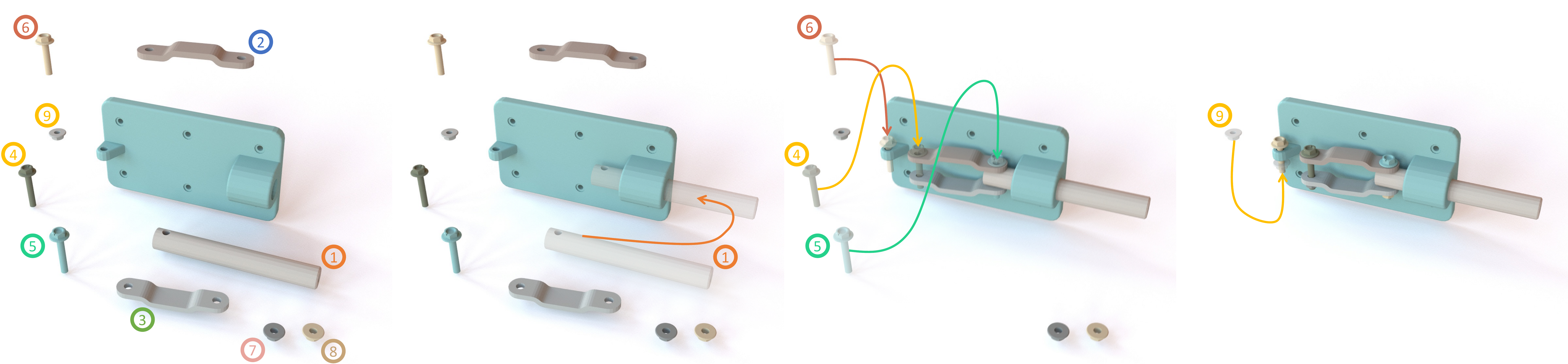}
    \vspace{-1mm}
    \caption{We present a physics-based planning algorithm for assembly tasks. Our algorithm follows an assembly-by-disassembly routine and utilizes a custom physics-based simulation to efficiently explore the assembly path space. Our algorithm efficiently determines the feasible assembly order (colored numbers) for a multi-part assembly and searches for a physically realistic assembly motion path (colored curves) for each assembly step. }
    \label{fig:teaser}
\end{teaserfigure}

\maketitle

\section{Introduction}
Real-world objects, such as vehicles, furniture, and electronic devices, are often complex assemblies made up of hundreds or thousands of parts. Each part is designed using computer aided design (CAD) software and then arranged into a digital assembly piece by piece. Considerations for physical assembly, known as Design for Assembly (DFA), are critical to ensure fast, efficient, and high-yield manufacturing. Likewise, considerations for physical disassembly, known as Design for Disassembly (DFD), are key to ensuring individual parts can be recycled or reused at end-of-life. The ability to automatically plan the steps required to assemble and disassemble designs created in CAD is an enabling technology for a number of downstream applications. In the design phase, DFA checks can be performed to correct potential issues before manufacture~\cite{melckenbeeck2020optimal}; assembly planning for high mix, low volume products can be greatly simplified with automatic generation of assembly instructions~\cite{agrawala2003designing}; and at end-of-life, robotic disassembly systems~\cite{ong2021product} can be quickly adapted to recycle product components. 

Despite significant research on assembly planning~\cite{santochi2002computer, ghandi2015review}, it remains a challenging problem for several reasons. Firstly, parts to be assembled can be of any shape --- even common parts, such as screws, contain complex surface geometry. Secondly, real-world assemblies often contain tightly packed parts, requiring assembly planning approaches that work well in a constrained space. Finally, the lack of large-scale assembly datasets has limited evaluation to small hand-picked collections. How well existing methods generalize across large-scale datasets remains unknown.

\begin{figure}
    \centering
    \includegraphics[width=0.45\textwidth]{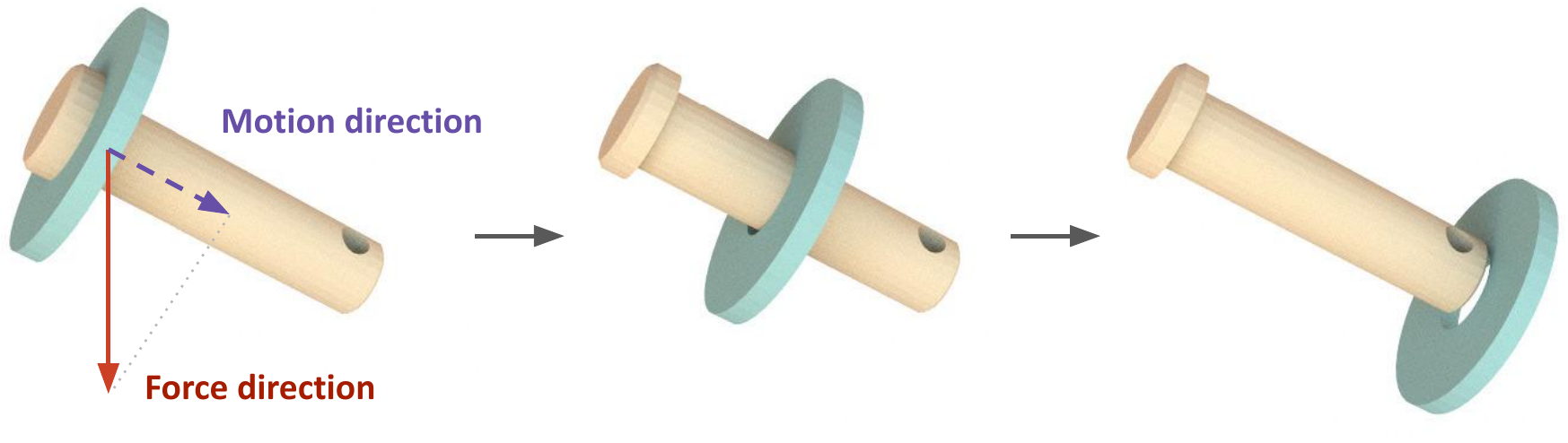}
    \vspace{-2mm}
    \caption{Our physics-based planning method is motivated by the observation that approximate force directions can be used for disassembly. In this example, the washer can be disassembled from the shaft as long as the force applied has a positive projection along the true disassembly direction, assuming no friction exists.}
    \vspace{-2mm}
    \label{fig:insight}
\end{figure}

To address this challenge, we introduce a novel physics-based method that takes an assembly-by-disassembly approach for sequence and motion planning with rigid assemblies.
Given a CAD model in an assembled state, we use a custom physics-based simulation to disassemble individual parts and recover viable assemble plans~(Figure~\ref{fig:teaser}). 
A key insight of our work is that physics-based simulation allows us to find disassembly motions using a discrete set of actions rather than exploring the full continuous 6D motion space. These discrete actions, represented as forces in canonical directions, can lead to successful motion when applied in physical simulation. For example, Figure~\ref{fig:insight} shows how a washer will slide along an inclined shaft even if the force is not applied directly along the disassembly direction.
Using this approach our custom physics-based simulation overcomes the difficulty of navigating parts through narrow passages with geometric motion planning, allowing us to explore highly constrained motion spaces without sampling invalid states in most situations.
To evaluate our approach and enable future research, we define a large-scale benchmark task for assembly planning consisting of thousands of physically valid assemblies --- two orders of magnitude larger than previous benchmarks. We make the following contributions in this work: 

\begin{itemize}
    \item We introduce an accurate and efficient physics-based simulator that is customized for assembly.
    \item We propose a novel physics-based assembly-by-disassembly planning method for translational and rotational assembly motion for arbitrary-shaped assemblies.
    \item We define a large-scale dataset and benchmark for assembly planning including thousands of physically valid assemblies.
    \item We evaluate our method on the full dataset and show a state-of-the-art success rate, computational efficiency, and generalization on various types of assemblies scaling to hundreds of parts.
\end{itemize}

To facilitate future assembly planning research, our code and dataset are available at \url{https://github.com/yunshengtian/Assemble-Them-All}.

\section{Related Work}

\paragraph{Assembly Sequence Planning (ASP)}

ASP arranges an optimal sequence in assembling all the components of a product and is a critical step for assembly system design~\cite{rashid2012review}. 
The ASP problem can be represented as a graph such as AND/OR graph~\cite{de1990and} to establish feasible assembly sequences. 
Different optimization techniques are applied to find the optimal assembly sequence among the feasible sequences~\cite{qin2007assembly, sinanouglu2005assembly, chen2006framework, chen2008three}.
The premise of these methods is that precedence constraints between parts need to be known such that feasible sequences could be discovered easily. Otherwise, the precedence relationships need to be annotated by domain experts or derived from the CAD models~\cite{su2009hierarchical, niu2003hierarchical}. 
Although many attempts have been made to reduce the search space, the number of potential assembly sequences can grow exponentially~\cite{ramos1998complexity} which makes identifying a feasible sequence impossible even for complex assemblies.

The idea of assembly-by-disassembly has been proposed as an important strategy for speeding up ASP, because assembled parts have better defined precedence and motion constraints than disassembled parts~\cite{demello1991}, reducing the search space. 
When all parts are rigid, a bijection exists between the assembly and disassembly sequences, meaning an assembly sequence can be obtained from the reverse order of its disassembly sequence with much less complexity.~\cite{ghandi2015review} 
The assembly-by-disassembly approach has been adopted in several applications in manufacturing and construction, including mechanical product assembly~\cite{demello1991, wang2014mechanical}, kit assembly~\cite{zakka2020form2fit}, and robotic additive construction~\cite{huang2021robotic}.
However, if the assembling direction are unknown or the assembling motions are non-linear, then more sophisticated assembly path planning methods are needed to make sequence planning feasible, even following an assembly-by-disassembly strategy.

\paragraph{Assembly Path Planning (APP)}

APP computes penetration-free paths for adding parts to a subassembly, given the initial and target poses of the parts.
Early works adopted exact geometric-reasoning methods that analyze part geometry to determine assembly directions~\cite{wilson1994geometric}. 
\citet{halperin2000general} presented a general framework for finding assembly motion called the motion space approach. Since geometric-reasoning approaches need to explicitly construct the Configuration space (C-space), they are computationally expensive for 3D space and non-linear motions. 

With the development of general-purpose path planning algorithm in robotics, sampling-based approaches have been proposed to solve problems with a large number of parts~\cite{masehian2021assembly}. Probabilistic Roadmap Method (PRM)~\cite{kavraki1996probabilistic} and Rapidly-exploring Random Tree (RRT)~\cite{lavalle1998rapidly} are the two notable sampling-based algorithms. 
PRM, RRT, and their variants
have been presented on assembly problems ~\cite{sundaram2001disassembly, le2009path, zhang2020c}. 
But they usually suffer from the `narrow passage' problem~\cite{hsu1999capturing}, especially in high-dimensional space where rotation needs to be considered.

Physics-based motion planning~\cite{zickler2009efficient} was developed as a competitive alternative to geometric-based motion planning 
and showed success with robotic navigation~\cite{sucan2011sampling} and manipulation~\cite{moll2017randomized}. However, physics-based motion planning has not been fully explored for assembly where the motion is much more constrained. 
In our work, we demonstrate the benefit of using physics-based planning for efficient exploration of assembly paths to overcome the `narrow passage' problem.

\paragraph{Physics-Based Simulation for Robotic Assembly}
With the ability to create a virtual and low-cost replica of the real world and provide a fast robotic evaluation platform, 
physics-based simulation 
has been widely used in control policy learning \cite{chen2021simple, 1606.01540, openai2020dexterous} and motion planning \cite{zickler2009efficient,sucan2011sampling,Zhou2014}. 

Despite its prevalence in robotics, simulating robotic assembly tasks is still far from perfect. Simulating contact-rich assembly tasks, such as a screw and nut in Figure~\ref{fig:sdf}, requires accurately representing the object surface while efficiently resolving contact constraints. Most existing simulators \cite{coumans2016pybullet, todorov2012mujoco, makoviychuk2021isaac} rely on a simplified collision proxy of the object shape for efficient collision detection and solve the linear complementary problem for contact handling. However, such collision proxies sacrifice the resolution of complex geometry shapes and is only suitable for simple assembly tasks, such as peg-in-hole \cite{hou2020data, chebotar2019closing} or box stacking \cite{aleotti2009efficient}. 
Another category of contact-rich simulators uses penalty-based contact models \cite{xu2021end, geilinger2020add}, but most only support collision detection between contact points and primitive shapes. Our simulation follows the line of penalty-based contact model, and is specialized for complex-shape assembly tasks via an extension of a signed-distance-field (SDF) collision detection. 
Concurrent to our work, \citet{narang2022factory} announced a GPU-based simulator for assembly tasks, which also utilizes the SDF shape representation.

\paragraph{Robotic Assembly}

Motion planning has been widely studied in robotic assembly for manufacturing~\cite{chen2021planning, wan2017assembly} and building construction~\cite{huang2021robotic, hartmann2021long}, albeit in less realistic scenarios where planners are limited to structured environments. Reinforcement Learning (RL)~\cite{sutton2018reinforcement}, on the other hand, has been applied to more industrial and general-purpose assembly tasks~\cite{thomas2018learning, yu2021roboassembly, de2021autonomous, fan2019learning} in unstructured environments~\cite{luo2021learning}. However, the trained policies are usually task-specific and struggle with arbitrary unseen objects. Our real-world dataset and SDF-based simulation
can aid RL-based methods in acquiring more generalized policies. Likewise, our physics-based planner can generate complex assembly motions that can directly guide robots or serve as demonstrations to learn from~\cite{zhu2018robot, roldan2019training}.


\section{Problem Formulation}

Given a collection of parts and their assembled state as input, our goal is to generate a sequence of physically realistic motion paths that position the parts into an assembled state, identical to the ground truth assembly.

Formally speaking, a state is defined as a vector $s$ that encodes the position of the part in the state space $S$. In our problem, $S$ can be either $\mathbb{R}^3$ for translational motion only or $SE(3)$ for both translational and rotational motion. 
For a given part $i$ in an assembly consisting of $M$ parts, a state $s^i$ is defined as a valid state as long as it does not have penetration with other parts, assuming other parts are fixed. A state $s^i$ is regarded as a disassembled state if the convex hull of the part geometry at $s^i$ does not have collision with the convex hull that encloses the geometries of all other parts. 
A disassembly path $P_D^i$ consists of a sequence of $n$ valid states $\{s_0^i,...,s_n^i\}$ that connects the assembled state $s_0^i$, given as input and a disassembled state $s_n^i$. Correspondingly, an assembly path $P_A^i=\{s_n^i,...,s_0^i\}$ exists.
Under these definitions, our problem aims at finding an ordered sequence of assembly paths for all $M$ parts that assemble them from scratch to the ground truth assembly with all the states being valid.

We make the following key assumptions in this problem: 1) Assemblies are all composed of rigid parts, which makes assembly-by-disassembly feasible; 2) Gravity or other force constraints and manipulation constraints are not considered as we mainly address the assembly planning from the motion level. In other words, we assume the assembly process is fully-actuated; 3) We assume parts can be completely assembled or disassembled sequentially, though handling of subassemblies could be extended by grouping and treating subassemblies as single parts.
\section{Physics-Based Assembly Planning}
Searching for an assembly strategy to move multiple parts from a disassembled state to a target assembled state is challenging, especially when the assembly process requires highly accurate insertion or screwing operations. This is because in a typical assembly task, the final assembled state space is much more constrained than the initial disassembled state space. As a result, an extremely high computational cost is incurred when searching from the less constrained space to a constrained final state. Motivated by this,
we solve the assembly problem in a reverse process, starting from the assembled state and searching for a disassembly solution, to reduce the overall search cost. We also leverage our custom physics-based simulator to further improve search efficiency. 

\begin{figure*}
    \centering
    \includegraphics[width=0.9\textwidth]{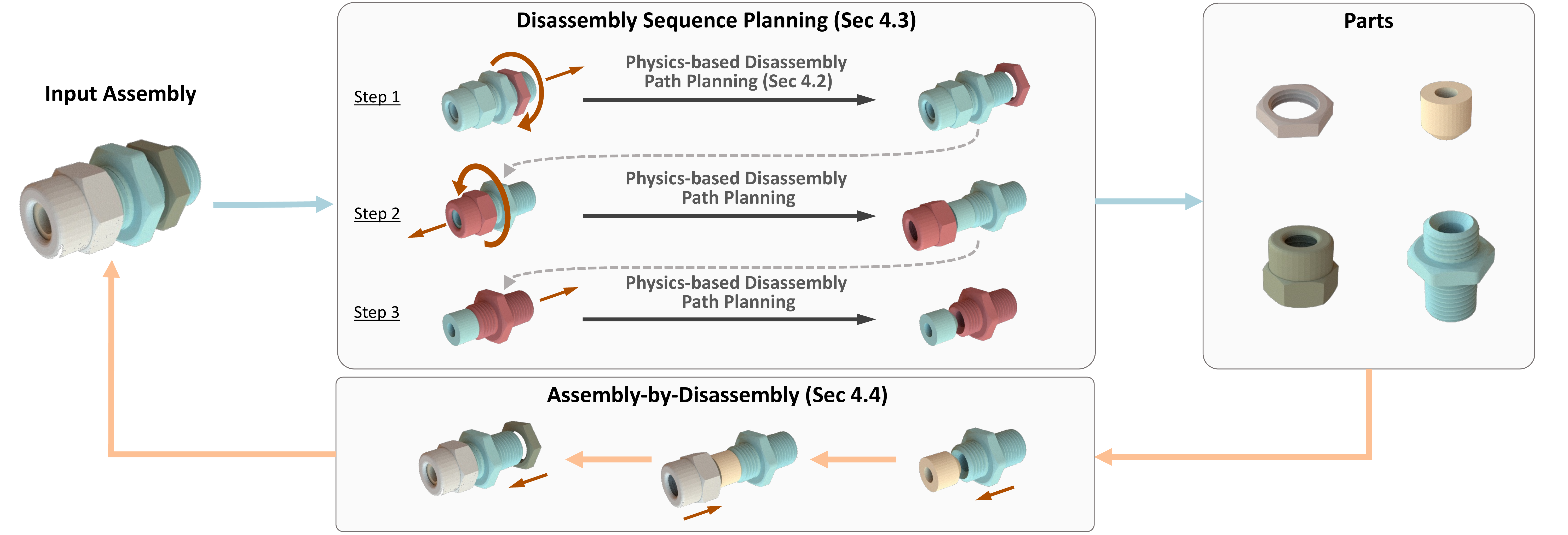}
    \vspace{-2mm}
    \caption{Overview of our physics-based assembly planning algorithm. Given the target assembled state as input (left), our algorithm finds the assembly strategy in two stages: disassembly and assembly. In the disassembly stage (upper center), our disassembly sequence planning algorithm searches for a feasible disassembly order for different parts, using a physics-based disassembly path planning algorithm that plans collision-free paths to disassemble each part. In the assembly stage (lower center), the algorithm reverses the disassembly path to obtain an assemble solution from individual parts (right).}
    \vspace{-2mm}
    \label{fig:algo_pipeline}
\end{figure*}

Our main algorithm pipeline is illustrated in Figure~\ref{fig:algo_pipeline}.
We first introduce our physics simulation for assembly in Section~\ref{sec:simulation} as the key component to our algorithm. 
Next, following the assembly-by-disassembly strategy, we illustrate our disassembly path planning algorithm for two-part disassembly in Section~\ref{sec:algo_path}, then our disassembly sequence planning algorithm for multi-part disassembly in Section~\ref{sec:algo_seq}. Finally, we present in Section~\ref{sec:algo_assembly} how to obtain the assembly sequence and path by reversing solutions from disassembly planning which completes our assembly-by-disassembly pipeline.

\subsection{Physics Simulation for Assembly}
\label{sec:simulation}
The existing robotics simulators \cite{coumans2016pybullet, todorov2012mujoco, makoviychuk2021isaac} primarily rely on the convex hull decomposition of the object shape for efficient collision detection, which is not suitable for simulating complex geometry shapes with high concavity. To reliably simulate contact-rich assembly motion for downstream assembly planning, we build our physics-based simulation upon the rigid body simulator developed by \citet{xu2021end} and extend it with the SDF representation of collision shapes to support complex collision geometries. 

Specifically, the simulator developed by \citet{xu2021end} is based on the reduced coordinate rigid body dynamics formulation of RedMax \cite{wang2019redmax}, and its dynamics equations are implicitly integrated in time with BDF1 scheme (backward Euler). To handle collisions, first the contacting parts are detected between each pair of collision shapes, where the mesh vertices of the first collision shape are sampled as contact points, and a distance function associated with the second collision shape is used to compute the following relevant collision information for each sampled contact point:
\begin{center}
\begin{tabular}{ll}
    $d$, \, $\dot{d}$ & \text{penetration distance and speed}\\
    $\vec{n}$           & \text{unit-length contact normal direction}\\
\end{tabular}
\end{center}
Then for each detected contact pair, the contact forces are computed by a penalty-based contact model as below:
\begin{align}
    \vec{f}_c = (-k_n + k_d \dot{d})d\vec{n}, \qquad
\end{align}
where $\vec{f}_c$ is the contact normal force on each contact pair, $k_n$ is the contact stiffness, and $k_d$ is the contact damping coefficient. In our work, we assume no friction exists as friction does not affect the assembly motion under the assumption that every part is rigid.
To efficiently acquire the required distance function, Xu et al.\shortcite{xu2021end} restrict one of each pair of collision shapes to be a primitive shape with an analytical distance function.

\subsubsection{Signed Distance Field Collision Detection}
\begin{figure}
    \centering
    \includegraphics[width=0.40\textwidth]{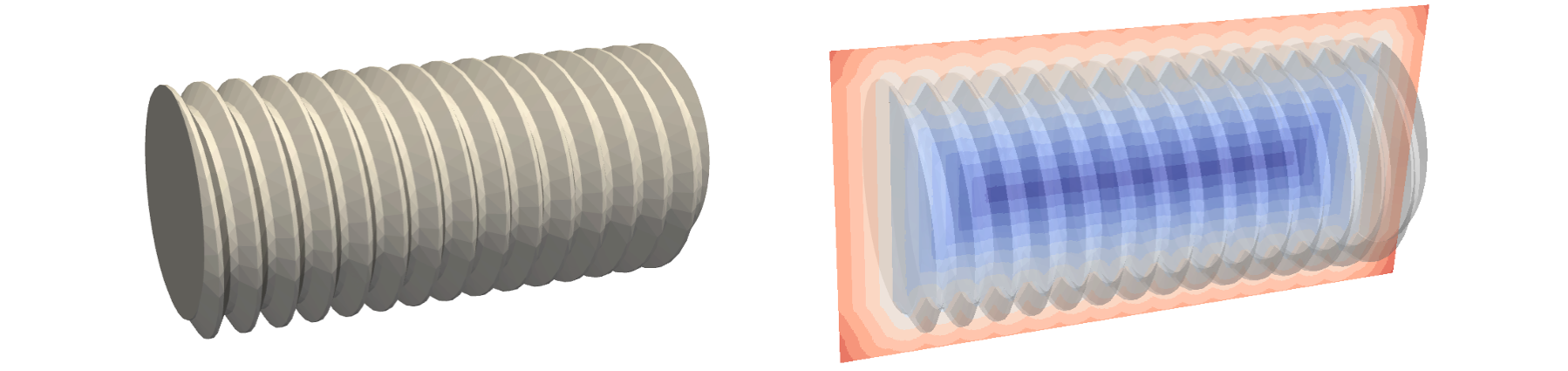}
    \vspace{-2mm}
    \caption{We construct a signed distance field (right) for each collision shape (left). The right figure shows the level set of the cross-section signed distance field of a fine-thread screw. }
    \vspace{-2mm}
    \label{fig:sdf}
\end{figure}
To support contact handling between complex assembly geometries (e.g., screw with fine threads), we equip each collision shape with an SDF to provide the distance function. Being an implicit surface representation for arbitrary shape geometry, an SDF is a function $g(\mathbf{x}): \mathbb{R}^3\rightarrow \mathbb{R}$ that maps a point in space to its closest distance to the represented shape surface (Figure \ref{fig:sdf}). The distance is negative when the point is inside the geometry while the distance is positive if the point is outside. 

Given the SDF of an object and the dynamics state $(\mathbf{x}, \dot{\mathbf{x}})$ of the contact point, the required collision information can be efficiently computed. The penetration distance $d = \min(g(\mathbf{x}), 0)$. The contact normal direction of the penetration is computed by the gradient of the SDF, i.e., $\vec{n} = \nabla g(\mathbf{x})$. The time derivative of the penetration distance is computed as $\dot{d} = \nabla g(\mathbf{x}) \cdot \dot{\mathbf{x}}$. And the relative tangential speed can be computed by projecting out the normal directional component from the relative speed between the two collision bodies $\vec{v}_t = \vec{v}_r(x) - \big(\vec{n}\cdot\vec{v}_r(x)\big)\vec{n}$, where $\vec{v}_r(x)$ is the relative speed of two collision bodies at the contact point $\textbf{x}$ which can be readily computed by rigid body simulator. For details on constructing the SDF grid, please refer to Appendix~\ref{supp:sdf}.

\subsection{Disassembly Path Planning}
\label{sec:algo_path}

For a given part in the assembly state $s_0$, we look for a sequence of states $\{s_0,...,s_n\}$ that removes the part from the rest of the assembly in a penetration-free manner. The state $s_n$ here is a valid disassembled state. 
Different from other path planning tasks, such as navigation or stacking objects, which usually operate in free space with infinite valid solution paths, disassembly requires highly constrained motion, similar to passing through a narrow corridor. This makes sampling-based geometric motion planning approaches less effective because of the low probability of obtaining a penetration-free state from random sampling. 
By contrast, humans rely on physics feedback to disassemble parts. As shown in Figure~\ref{fig:insight}, even though we may apply forces that are not perfectly aligned with the ground-truth disassembly motion, we are able to infer the correct motion direction after the induced movement of the part is observed.

From this key observation, we propose a physics-based planner that efficiently plans the disassembly motion by leveraging feedback from physics. 
We formulate the disassembly path planning as a tree search problem, which starts from the assembled state and searches for a sequence of actions until a disassembled state has been found or some time/depth limitation of the search has been reached. The outline of our algorithm is illustrated in Algorithm~\ref{alg:path_plan}. 

Specifically, we adopt breadth-first search (BFS) and maintain a queue $Q$ to store the states to be searched from. The search starts with the queue only containing the assembled state $s_0$. In each iteration, we dequeue a state $s$ from $Q$, and loop over all actions $a_i$ in a pre-defined action space $A$ to generate child states of the disassembly tree. In our method, the action space consists of unit forces or torques of the given state space. More specifically, in translational-motion-only cases, $A=\{[\pm 1,0,0],[0,\pm 1,0],[0,0,\pm 1]\}$ which has 6 actions corresponding to 6 translational degrees of freedom (DoF); similarly, when rotational motion is enabled, $A$ has 12 actions for 6 translational DoF and 6 rotational DoF. For each action $a_i \in A$, we continuously apply this action starting from the dequeued state $s$ in physics-based simulation with a certain time step $\Delta t$, until the new state $s_i$ becomes either a disassembled state or a similar state to one that has been searched in the past. If $s_i$ is a disassembled state, then the disassembly path planning succeeds and we can obtain the path from the initial assembled state $s_0$ to $s_i$ as a disassembly path which can be retrieved from backtracking the search tree from $s_i$. Otherwise, if $s_i$ is detected to be a similar state to any previous state,  
we stop applying this action, enqueue the current state $s_i$, and continue searching with the rest of actions. Here we measure the similarity between two states by their translational and rotational distances. The translational distance is defined by the Euclidean distance $\lVert s_{a_t} - s_{b_t} \rVert$ between two states, where $s_{a_t}$ and $s_{b_t}$ are the linear components of the two states. The rotational distance is defined by the geodesic distance $\lVert\ln(s_{a_r}^{-1}s_{b_r})\rVert$ between the quaternions of two states, where $s_{a_r}$ and $s_{b_r}$ are the quaternion representations of the angular components of two states. Two states are similar if their translational and rotational distances are within thresholds $\delta_t$ and $\delta_r$ respectively.

As a result, BFS continuously explores the state space in a hierarchical manner and we enforce it to always explore novel states by checking state similarity. Note that the dynamics of our physics-based simulation constrains BFS to search in a physically valid subspace rather than the full state space which avoids the narrow passage problem in previous geometric motion planning methods. BFS is empirically quite effective because most real-world assemblies are designed to be assembled/disassembled within only one or a few sequential actions, which means that the depth of BFS is usually small. 

Our BFS-guided disassembly path planning method works well for general industrial assemblies as empirically proved by our evaluation. But for rare cases like disassembling parts from a long zig-zag path, BFS might be less efficient. In this case, we suggest a more sophisticated search algorithm such as Monte-Carlo Tree Search (MCTS)~\cite{coulom2006efficient} along with carefully designed heuristics might be applied to speed up the search, though it might not perform equally well on simpler assemblies.

\begin{algorithm}[t]
\SetAlgoLined
\KwIn{Assembled state $s_0$, timeout $t_{\max}$, max BFS depth $d_{\max}$.}
\KwOut{A disassembly path $P_D = \{s_0,...,s_n\}$.}
$Q$ = EmptyQueue()\;
$Q$.Enqueue($s_0$)\;
\While{time $t < t_{\max}$ and BFS depth $d \leq d_{\max}$}{
    $s$ = $Q$.Dequeue()\;
    \For{$a_i$ in action space $A$}{
        $s_i$ = $s$\;
        \Do{$s_i$ is not similar to any other past states}{
            $s_i$ = Simulate($s_i$, $a_i$, $\Delta t$)\;
            \If{IsDisassembled($s_i$)}{
                \Return{GetPath($s_0$, $s_i$)}\;
            }
        }
        $Q$.Enqueue($s_i$)\;
    }
}
\Return{failed}\;
\caption{Disassembly path planning guided by BFS}
\label{alg:path_plan}
\end{algorithm}

\subsection{Disassembly Sequence Planning}
\label{sec:algo_seq}

Because of the internal precedence relationship between parts, most real world assemblies consist of multiple parts and require specific sequences to assemble/diassemble them. However, the precedence relationship is usually unknown beforehand.
Without a carefully designed algorithm, the time complexity to figure out the correct disassembly sequence may grow up quadratically as the number of parts in an assembly increases. When coupled with the path planning algorithm, the overall running time for multi-part disassembly
algorithm could be extremely slow ~\cite{ebinger2018general}. 

We propose a method to efficiently plan the disassembly sequence by progressively expanded BFS, see Algorithm~\ref{alg:seq_plan_disassembly}. Given the assembled states $\textbf{s} = \{s_0^1,...,s_0^M\}$ of all $M$ parts, we look for an ordered sequence of disassembly paths 
$\textbf{P}_D = \{\{s_0^{I_1},...,s_{n}^{I_1}\},...,\{s_0^{I_M},...,s_{n}^{I_M}\}\}$ 
that connect the assembled state and an disassembled state for each part and satisfy precedence relationships. Here, $I_1,...,I_M$ correspond to ordered indices of the part sequence ranging from 1 to $M$. In each iteration, our sequence planner tries to disassemble each of the remaining parts of the assembly by our path planner from Algorithm~\ref{alg:path_plan}. If any part is successfully disassembled, we append its disassembly path to the sequence $\textbf{P}_D$. We repeat this procedure until all parts are disassembled or we reach the timeout. 

Executing this procedure naively without limiting the BFS depth could be time-consuming due to the excessive time wasted on trying to disassemble parts that are blocked by other parts. Therefore, we take a progressive approach that starts with a shallow BFS and gradually increases the depth of BFS as the sequence planning progresses. For example, in the first iteration, we limit our path planning algorithm for disassembling all parts to have a maximal BFS depth $d_{\max} = 1$. Then, for each part it tries to disassemble, if our path planning algorithm fails to find a disassembly path with a single step of action application, we will temporarily stop trying this part and move on to other parts. There could be two reasons for this failure: 1) this part is blocked by other parts, i.e., conflict with the precedence relationships; 2) this part is not blocked but requires multiple actions to disassemble. Our progressive approach prevents us from wasting 
unnecessary effort in the first case while allowing us to try more actions in the later stage to handle the second case. 
Again, since most real-world assemblies are designed to be easily assembled/disassembled within one or a few actions, the first case is much more common and by early termination of the failed attempts we empirically observed a significant speed-up 
compared to working with unlimited BFS depth in Section~\ref{sec:eval_multipart}.

\begin{algorithm}[t]
\SetAlgoLined
\KwIn{Assembled states $\textbf{s} = \{s_0^1,...,s_0^M\}$ of all $M$ parts, path planning timeout $t_{\max}$, sequence planning timeout $T_{\max}$.}
\KwOut{An ordered sequence of disassembly paths $\textbf{P}_D = \{P_D^{I_1},...,P_D^{I_M}\} = \{\{s_0^{I_1},...,s_{n}^{I_1}\},...,\{s_0^{I_M},...,s_{n}^{I_M}\}\}$.}
$\textbf{P}_D = \{\}$, max BFS depth $d_{\max}$ = 1\;
\While{time $t < T_{\max}$}{
    \For{part $i$ in all remaining parts of the assembly}{
        $P_D^i$ = DisassemblyPathPlanning($s_0^i$, $t_{\max}$, $d_{\max}$)\;
        \If{$P_D^i \neq$ failed}{
            $\textbf{P}_D$.Append($P_D^i$)\;
            Remove part $i$ from the assembly\;
        }
    }
    \If{all $M$ parts are disassembled}{
        \Return{$\textbf{P}_D$}\;
    }
    $d_{\max} = d_{\max} \mathrel{+} 1$\;
}
\Return{failed}\;
\caption{Disassembly sequence planning with Prog. BFS}
\label{alg:seq_plan_disassembly}
\end{algorithm}

\subsection{Assembly-by-Diassembly}
\label{sec:algo_assembly}

After the disassembly sequence and paths have been found, we can reverse them and connect with the initial states to construct the entire assembly sequence and paths through assembly-by-disassembly strategy,
as illustrated in Algorithm~\ref{alg:seq_plan_assembly}. To simplify the notation we ignore the superscript $I_i$ for part index. Each disassembly path $P_D$ obtained from Algorithm~\ref{alg:path_plan} and \ref{alg:seq_plan_disassembly} connects the assembled state $s_\text{goal}$ and a disassembled state $s_\text{dis}$, but the path between $s_\text{dis}$ and initial state $s_\text{init}$ (which is specified by the user) is still missing. Different from the disassembly path, this path does not require highly constrained motion since both states are disassembled, we can easily apply a standard geometric motion planning algorithm such as RRT-Connect~\cite{kuffner2000rrt} to efficiently plan a collision-free path $P_C$ between $s_\text{init}$ and $s_\text{dis}$. Therefore, we can obtain the assembly path $P_A$ from initial state $s_\text{init}$ to assembled state $s_\text{goal}$ by concatenating $P_C$ and the reversed disassembly path $P_D$. By looping over the disassembly sequence reversely, we finally obtain an ordered assembly sequence containing all the assembly paths.

\begin{algorithm}[h]
\SetAlgoLined
\KwIn{Initial states $\textbf{s}_\text{init} = \{s_{\text{init}}^1,...,s_{\text{init}}^M\}$ and assembled states $\textbf{s}_\text{goal} = \{s_\text{goal}^1,...,s_\text{goal}^M\}$ of all $M$ parts.}
\KwOut{An ordered sequence of assembly paths $\textbf{P}_A = \{P_A^{I_M},...,P_A^{I_1}\} = \{\{s_\text{init}^{I_M},...,s_\text{goal}^{I_M}\},...,\{s_\text{init}^{I_1},...,s_\text{goal}^{I_1}\}\}$.}
Compute ordered disassembly paths $\textbf{P}_D$ from Alg.~\ref{alg:seq_plan_disassembly} given $\textbf{s}_\text{goal}$\;
$\textbf{P}_A = \{\}$\;
\For{$i$ in $M,...,1$}{
    Obtain the disassembled state $s_\text{dis}^{I_i}$ from disassembly path $P_D^{I_i}$\;
    Compute a collision-free path $P_C^{I_i}$ connecting $s_\text{init}^{I_i}$ and $s_\text{dis}^{I_i}$ by a standard motion planning algorithm (e.g., RRT-Connect)\;
    Connect assembly path $P_A^{I_i} = P_C^{I_i} + \text{Reverse}(P_D^{I_i})$\;
    $\textbf{P}_A$.Append($P_A^{I_i}$)\;
}
\Return{$\textbf{P}_A$}\;
\caption{Assembly by disassembly}
\label{alg:seq_plan_assembly}
\end{algorithm}

\section{Assembly Dataset}
\label{sec:dataset}
To accurately evaluate our method we define a large-scale dataset for assembly planning.
A focus of popular large-scale 3D shape datasets has been on synthetic assemblies that are semantically segmented~\cite{chang2015shapenet, mo2019partnet} with part mobility labels~\cite{xiang2020sapien, wang2019shape2motion, yanRPMNet19, hu2017learning}. In comparison, real world CAD assemblies contain detailed design, such as fastener stacks, and are segmented using different criteria, such as manufacturing process and ease of repair. We build upon the recent release of several realistic CAD assembly datasets~\cite{koch2019abc, willis2021fusion, willis2022joinable, jones2021automate} to define our dataset and benchmark for assembly planning. To the best of our knowledge, there does not exist a large-scale open-source dataset for assembly planning research.

\begin{figure}
    \centering
    \includegraphics[width=0.48\textwidth]{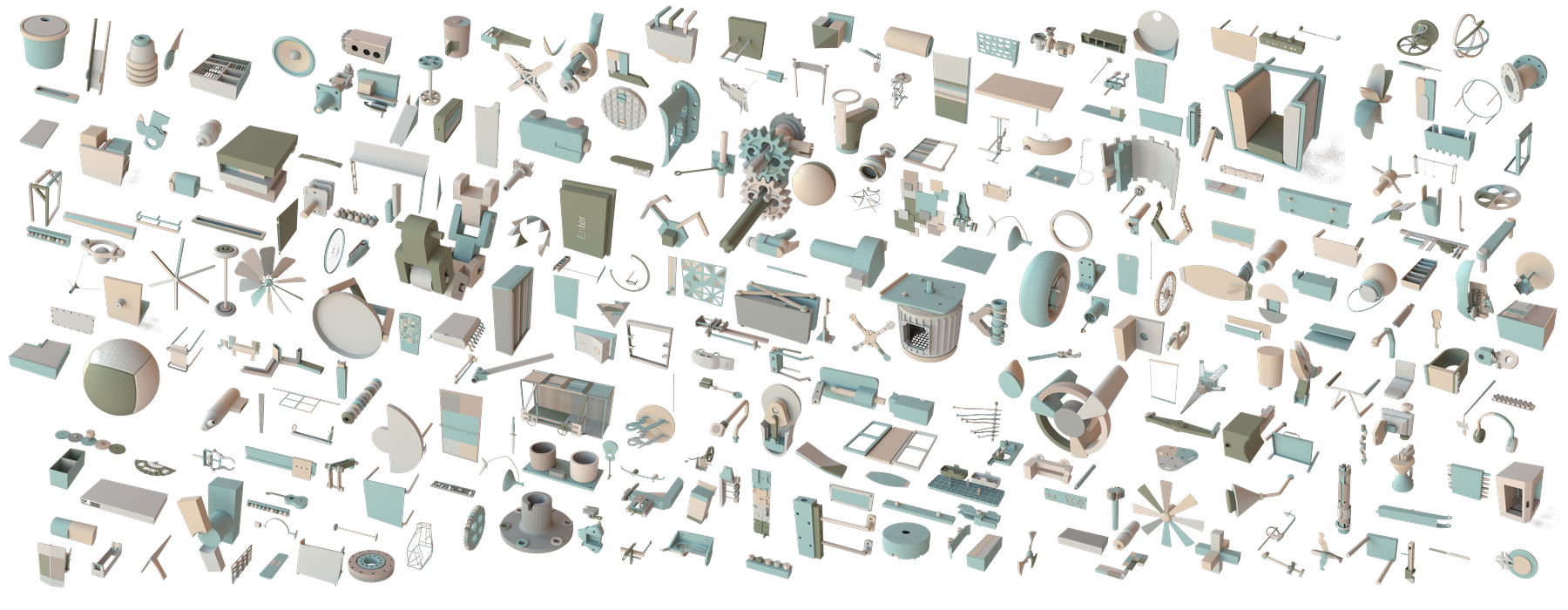}
    \vspace{-1.5em}
    \caption{Overview of our dataset containing thousands of physically valid assemblies suitable for use with assembly planning tasks.}
    \vspace{-2mm}
    \label{fig:dataset}
\end{figure}

We derive our dataset from assemblies created in Autodesk Fusion 360 and provided by~\citet{willis2022joinable} as well as assemblies from \citet{ebinger2018general} and \citet{zhang2020c}. To adapt the data for use with assembly planning we perform several geometric pre-processing steps to produce watertight, unique, and normalized meshes suitable for accurate collision checking. We filter out assemblies where parts are in a non-assembled state, or in a state of overlap.
Table~\ref{tab:dataset_stats} lists the total number of assemblies and their distribution by number of parts in each assembly. 
Figure~\ref{fig:dataset} shows a random sampling of assemblies in the dataset.
For a more detailed description of the geometric pre-processing steps on the dataset, please refer to Appendix~\ref{supp:data_process}.

\begin{table}[t]%
\small
\vspace{-0.5em}
\caption{Dataset statistics.}
\vspace{-1em}
\label{tab:dataset_stats}
\begin{center}
\begin{tabular}{rccccc}
  \toprule
  \textbf{\# Parts} & 2 & 3-9 & 10-49 & 50-235 & \textbf{Total} \\
  \textbf{\# Assemblies} & 8776 & 2620 & 1468 & 106 & \textbf{12970} \\
  \bottomrule
\end{tabular}
\end{center}
\end{table}%

\section{Evaluation}
\label{sec:evaluation}

In this section, we perform experiments to evaluate our method on the proposed dataset with thousands of assemblies to demonstrate the advantage of our physics-based planning approach in terms of success rate and computational efficiency. For fairness, all baseline methods also follow the assembly-by-disassembly approach.
Finally in Section~\ref{sec:eval_interlocking}, we show a naive extension of our method to solve more complicated assemblies that involve simultaneous movements of parts to disassemble.

We run all our experiments on Amazon EC2 instances (c5.24xlarge) with 96 vCPUs and 192G memory. The detailed hyper-parameter settings of all the methods can be found in Appendix~\ref{supp:hp}.

\subsection{Baseline Methods}
\label{sec:eval_baseline}

We choose the following path planning methods which are the most relevant baselines to compare:
\textbf{RRT}: Rapidly-Exploring Random Tree~\cite{lavalle1998rapidly} is one of the most standard geometric path planning approaches, which requires an explicitly specified goal state.
In our expriments, we use a randomly selected disassembled state as the goal for RRT.
\textbf{T-RRT}: Targetless-RRT~\cite{aguinaga2008parallel} is a modified version of RRT that is specifically designed for disassembly planning, which does not require a specified specific goal state. 
\textbf{MV+T-RRT}: Mating Vector + Targetless-RRT, a hybrid approach proposed in \cite{ebinger2018general} that first uses face normals of objects as "mating vectors" to try a straight-line disassembly, then if it fails, T-RRT is applied. This is the state-of-the-art approach in disassembly path planning to the best of our knowledge.
\textbf{BK-RRT}: Behavioral Kinodynamic
Rapidly-Exploring Random Trees~\cite{zickler2009efficient} is a classic physics-based planning algorithm similar to our method that uses a physics-based simulation to randomly explore new states. BK-RRT is shown to be successful in navigation and manipulation tasks but has not been evaluated in assembly or disassembly.
More details of the baseline methods can be found in Appendix~\ref{supp:algo}. 

\subsection{Two-Part Assembly}
\label{sec:eval_twopart}

We evaluate our path planning algorithm and baseline approaches on the two-part dataset and report the performance comparison in Table~\ref{tab:twopart_success} (left). 
The reported success rates are calculated based on a 300s timeout, averaged across results from 6 random seeds.
To further validate that our method generalizes to non-axis-aligned disassembly directions, we randomly rotate all the assemblies in the two-part dataset and report performance in Table~\ref{tab:twopart_success} (bottom).

\begin{figure}
    \centering
    \includegraphics[width=0.45\textwidth]{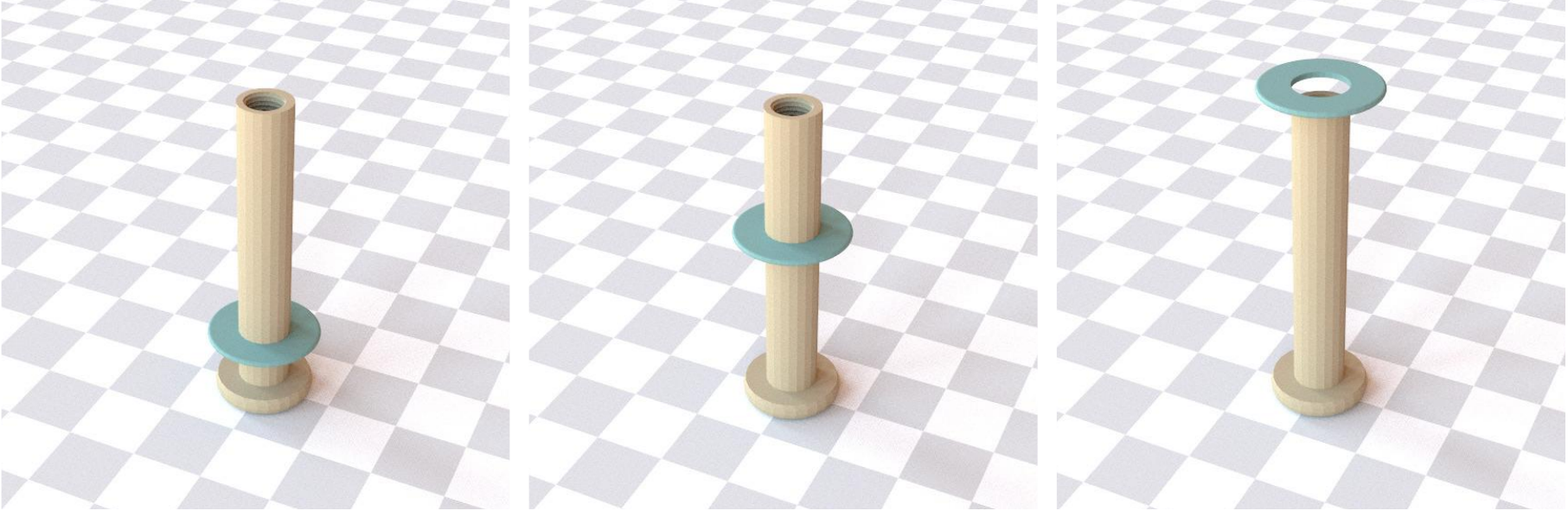}
    \caption{A tightly-fitted two-part disassembly (left: assembled, right: disassembled). This can be easily solved by our method while being challenging for baseline methods due to the long and narrow passage.}
    \label{fig:twopart_motion}
    \vspace{-1em}
\end{figure}

On this dataset, our approach achieves almost $100\%$ success rate which is robust to various shapes and disassembly motions. Geometric baseline methods (RRT, T-RRT and MV+T-RRT) show decent success rates but would fail when disassembly motion is too constrained such as the task shown in Figure \ref{fig:twopart_motion}. In this example, the collision-free disassembly path should be strictly aligned with the axis of the cylinder, which is unlikely to be sampled from the whole motion space without assistance of a physics simulator. 
In contrast, both our approach and the physics-based baseline (BK-RRT) utilizes the physics simulator to mitigate the narrow passage problem. However, BK-RRT still fails in the task shown in Figure \ref{fig:twopart_motion}, that is because the random states explored by BK-RRT make it hard to keep moving in a single direction for a long period, thus result in a oscillatory motion of the ring on the cylinder. In our approach, the usage of BFS and state similarity check helps move part along one direction much easier.
Another typical failure of the baseline methods is the difficulty of dealing with assemblies that require non-trivial rotational motions such as screws, as shown in Figure~\ref{fig:rotation_motion}. For visualization on more results, please refer to Appendix~\ref{supp:results}.

\begin{table}%
\small
\caption{Success rate (\%) comparison on the entire two-part assembly dataset along with specific comparison on different rotational assemblies. Our method reaches almost 100\% success rate on the two-part dataset and outperforms baseline methods by a large margin on rotational assemblies. The bottom row shows the results of our method on the same dataset with random rotations applied to the data.}
\label{tab:twopart_success}
\begin{center}
\begin{tabular}{l|c|cccc}

  \toprule
  \textbf{Method} & \textbf{Two-Part} & \multicolumn{4}{c}{\textbf{Rotational}} \\
  {} & Overall & Screw & Puzzle & Others & Overall \\
  \midrule
  RRT       & 84.5 & 0.0 & 20.8 & 0.0 & 6.9 \\
  T-RRT     & 97.4 & 2.1 & 18.8 & 0.0 & 6.9 \\
  MV+T-RRT     & 97.8 & 12.5 & 18.8 & 0.0 & 10.4 \\
  BK-RRT    & 93.7 & 12.5 & 62.5 & \textbf{81.3} & 52.1 \\
  \textbf{Ours} & \textbf{99.8} & \textbf{75.0} & \textbf{87.5} & 50.0 & \textbf{70.8} \\
  \midrule
  \textbf{Ours} (Rotated) & 99.8 & 37.5 & 62.5 & 87.5 & 62.5 \\
  \bottomrule

\end{tabular}
\end{center}
\end{table}%

\subsubsection{Rotational Assembly}
\label{sec:eval_rotation}

In this section, we specifically evaluate all algorithms on rotational assemblies, which are notoriously difficult for existing path planning approaches. In order to showcase different complexities of rotations, we prepare a rotational assembly dataset that consists of three categories of rotation to evaluate algorithm performance: 1) \textbf{Screw} for screws and nuts which are the most common rotational components in industrial assemblies; 2) \textbf{Puzzle} refers to puzzle assemblies that are mostly for entertainment purposes but are difficult to disassemble even for humans, gathered from \cite{zhang2020c}; 3) \textbf{Others} for all other types of rotations, e.g., removing a ring along a curved tube. The rotational dataset has 24 assemblies in total (8 in each category) due to the scarcity of rotational assemblies.
A complete overview of the rotational dataset can be found in Appendix~\ref{supp:rot_data}.

\begin{figure}
    \centering
    \includegraphics[width=0.45\textwidth]{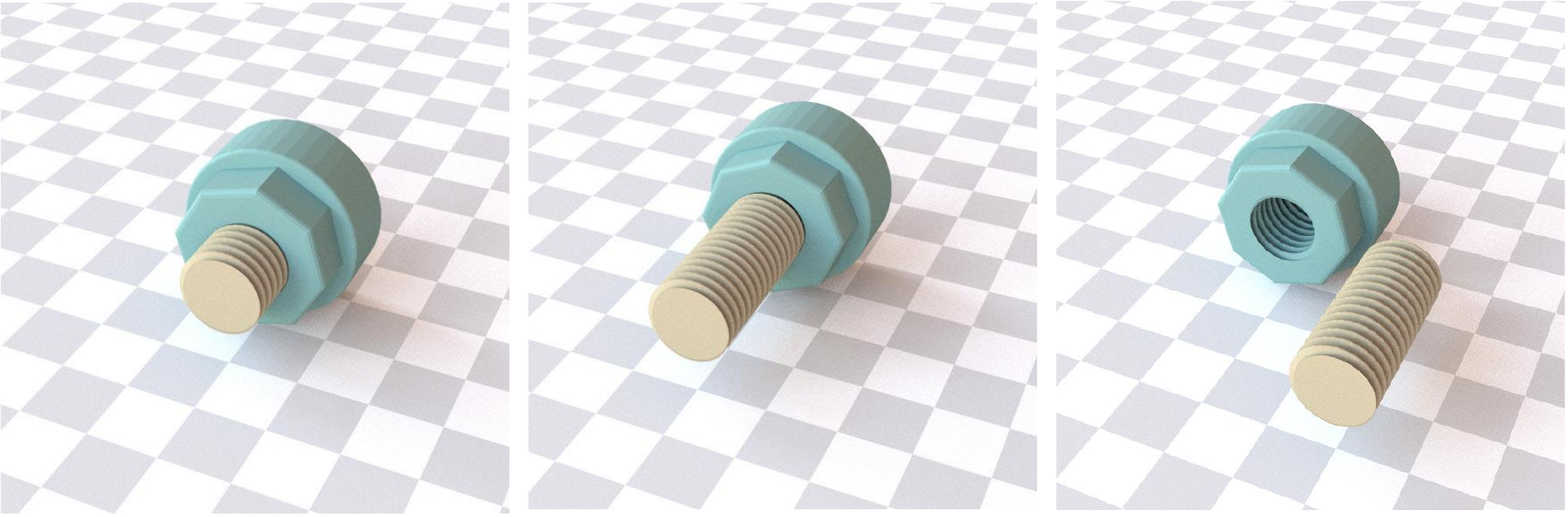}
    \caption{Screw disassembly (left: assembled, right: disassembled). Our method successfully generates a long sequence of screwing motions and a short translation at the end to disassemble the screw from the nut.}
    \label{fig:rotation_motion}
    \vspace{-1em}
\end{figure}

We report the results in Table ~\ref{tab:twopart_success} (right). The reported success rates are calculated based on a 600s timeout, averaged across results from 6 random seeds. From the results, we observe significantly better success rates of physics-based methods (i.e. BK-RRT and ours) in rotational disassembly path planning in all three categories of rotations.  Our method outperforms other baselines by a large margin in \textbf{Screw} and \textbf{Puzzle} assemblies while being worse than BK-RRT in the \textbf{Others} category. 
Figure ~\ref{fig:rotation_motion} shows a screw example which can be successfully solved by our approach but fails on all other baseline methods.

For puzzle assemblies, \citet{zhang2020c} proposed a sampling-based geometric planning approach that combines with a learning-based geometric feature extractor to guide exploration. Their method takes hours of computation to find a disassembly path while we achieve a comparable success rate within several minutes on a subset of their puzzles. However, their approach is more specialized for solving more complicated puzzles and it is not designed for general types of assemblies.

\subsection{Multi-Part Assembly}
\label{sec:eval_multipart}
We evaluate our sequence planning algorithm on the portion of dataset with multi-part assemblies. We partition the multi-part dataset based on the assembly size into \textbf{Small}, \textbf{Medium} and \textbf{Large} categories corresponding to assemblies of 3-9, 10-49 and 50+ parts respectively. 
The statistics of each category are shown in Table~\ref{tab:dataset_stats}.
For baseline methods, we apply a similar disassembly sequence planning procedure as Algorithm~\ref{alg:seq_plan_disassembly} but without progressive BFS. For our method, we compare two variants: \textbf{Prog. BFS} as sequence planning with progressive BFS illustrated in Algorithm~\ref{alg:seq_plan_disassembly} and \textbf{Full BFS} as Algorithm~\ref{alg:seq_plan_disassembly} without limiting the max BFS depth. 

We first compare the success rates of different algorithms in Table \ref{tab:multipart_success}. The results are produced with a 120s timeout for each path planning attempt and a 7200s timeout for the entire sequence planning, averaged from 3 random seeds. Table~\ref{tab:multipart_success} shows that our method (either Full BFS or Prog. BFS) outperforms other baselines on all different assembly sizes. 

\begin{figure}
    \centering
    \includegraphics[width=0.45\textwidth]{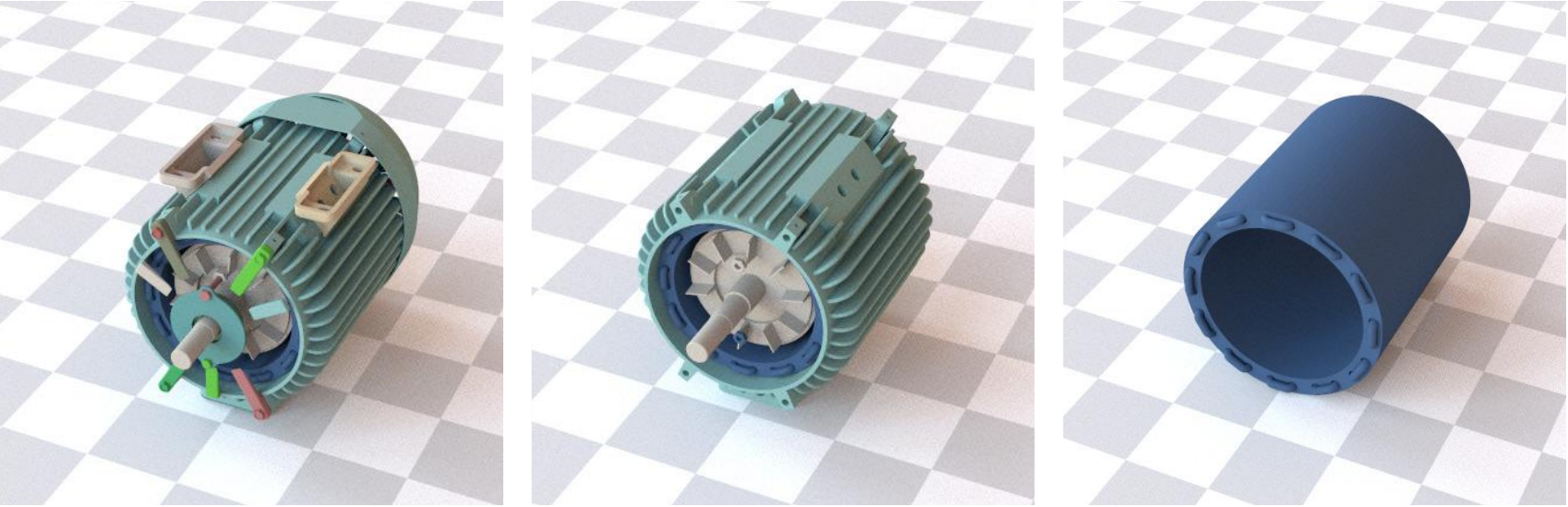}
    \caption{53-part engine disassembly (left: 53 parts left, middle: 11 parts left, right: 4 parts left). Our method with progressive BFS disassembles the engine completely within 5 minutes, an order of magnitude faster than current state-of-the-art approaches.}
    \label{fig:multipart_motion}
    \vspace{-1em}
\end{figure}

Furthermore, our method is capable of disassmbling complex many-part assemblies much faster than existing state-of-the-art approaches. For example, we successfully disassemble the entire 53-part Engine example (Figure \ref{fig:multipart_motion}) from \cite{ebinger2018general} in 5 minutes while their MV+T-RRT takes hours to complete. We also empirically demonstrate in Table~\ref{tab:multipart_time} that our sequence planning method guided by progressive BFS effectively cuts down the computation time by 3.5x on average compared to using the full BFS.

\begin{table}%
\small
\caption{Success rate comparison (\%) on multi-part assembly dataset. Our method consistently outperform baseline methods on all sizes of assemblies.}
\label{tab:multipart_success}
\begin{center}
\begin{tabular}{l|cccc}
  \toprule
  \textbf{Method} & \multicolumn{4}{c}{\textbf{Multi-Part}} \\
  {} & Small & Medium & Large & Overall\\
  \midrule
  RRT & 90.0 & 78.1 & 44.3 & 84.7 \\
  T-RRT & 94.0 & 84.8 & 52.5 & 89.7 \\  
  MV+T-RRT & 94.2 & 85.2 & 53.8 & 90.0 \\
  BK-RRT & 96.4 & 91.6 & 51.9 & 93.6 \\
  \textbf{Ours (Full BFS)} & \textbf{99.1} & 96.8 & 71.1 & 97.6 \\
  \textbf{Ours (Prog. BFS)} & 99.0 & \textbf{97.3} & \textbf{76.7} & \textbf{97.9} \\
  \bottomrule
\end{tabular}
\end{center}
\end{table}%

\begin{table}%
\small
\caption{Mean disassembly time per part (s) comparison on multi-part assembly dataset with different sizes of assemblies. Our proposed Prog. BFS gives large computational efficiency improvement compared to Full BFS.
}
\label{tab:multipart_time}
\begin{center}
\begin{tabular}{l|cccc}
  \toprule
  {} & Small & Medium & Large & Overall\\
  \midrule
  Full BFS & 18.5 & 38.4 & 30.6 & 25.6 \\
  Prog. BFS & 4.5 & 10.1 & 19.4 & 6.7 \\
  \bottomrule
\end{tabular}
\end{center}
\end{table}%

\subsection{Sub-Assembly and Interlocking Assembly}
\label{sec:eval_interlocking}

Although most data in our dataset can be sequentially assembled, in this section, we extend our method to handle assemblies requiring simultaneous movements of multiple parts to disassemble. The first scenario is the use of sub-assemblies, where multiple parts form a group that must be disassembled together and follow the same disassembly motion. The second scenario is interlocking assemblies where multiple parts must follow different motions to free the assembly from the interlocked state. Here we describe a naive extension of our physics-based planning method to handle sub-assembly and interlocking assembly: 

Given an $M$-part assembly, assuming at least $m$ parts ($1 < m < M$) need to be moved to disassemble it, we need to figure out which $m$ parts to move (sequence planning) and how these parts should move (path planning). For sequence planning, different from Algorithm~\ref{alg:seq_plan_disassembly} where it searches for a single part to disassemble, we extend it to search for all $m$-part combinations. For path planning, sub-assemblies can be treated as single parts and follow the same procedure in Algorithm~\ref{alg:path_plan}, but for interlocking assemblies, instead of searching for single actions for single parts, we need to search for $m$ actions to be applied to the $m$ parts respectively. 

\begin{figure}
    \centering
    \includegraphics[width=0.48\textwidth]{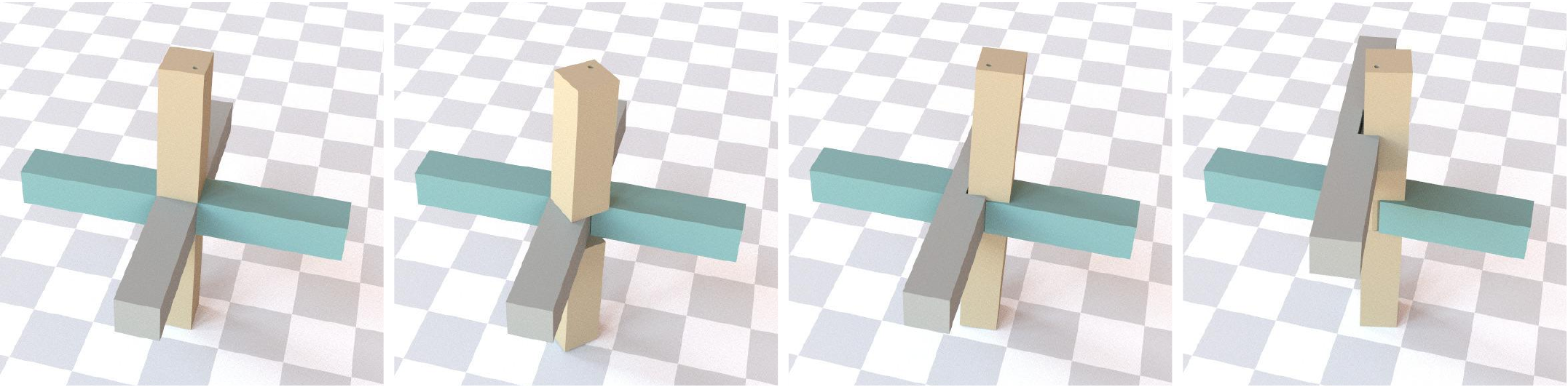}
    \caption{A 3-part interlocking assembly disassembled by the movement of 2 parts (left: assembled, right: disassembled). Our extended physics-based planner discovers the correct disassembly strategy by simultaneously applying a counterclockwise rotational torque on the beige block and an upward vertical force on the grey block.}
    \label{fig:interlocking}
\end{figure}

Through such an exhaustive search combined with physics-based planning, Figure~\ref{fig:interlocking} shows an example where our extended method identifies the correct parts to disassemble along with the correct actions to apply. This is achieved by simultaneously applying a rotational torque around the z-axis (vertical axis) to the beige block and a positive translational force along the z-axis to the grey block. Initially, the translational force applied on the grey block will be counteracted by the contact force from the beige block, but after the beige block rotates for 180\degree, it immediately pulls the grey block upwards to disassemble. In this case, the physical interactions between blocks are crucial to constrain and guide the disassembly motion towards the correct direction.

\section{Limitations}
\label{sec:failure}

In this section, we analyze the main failure cases of our method from the experimental results shown in Section~\ref{sec:evaluation}.

\begin{figure}
    \vspace{-1.0em}
    \centering
    \includegraphics[width=0.45\textwidth]{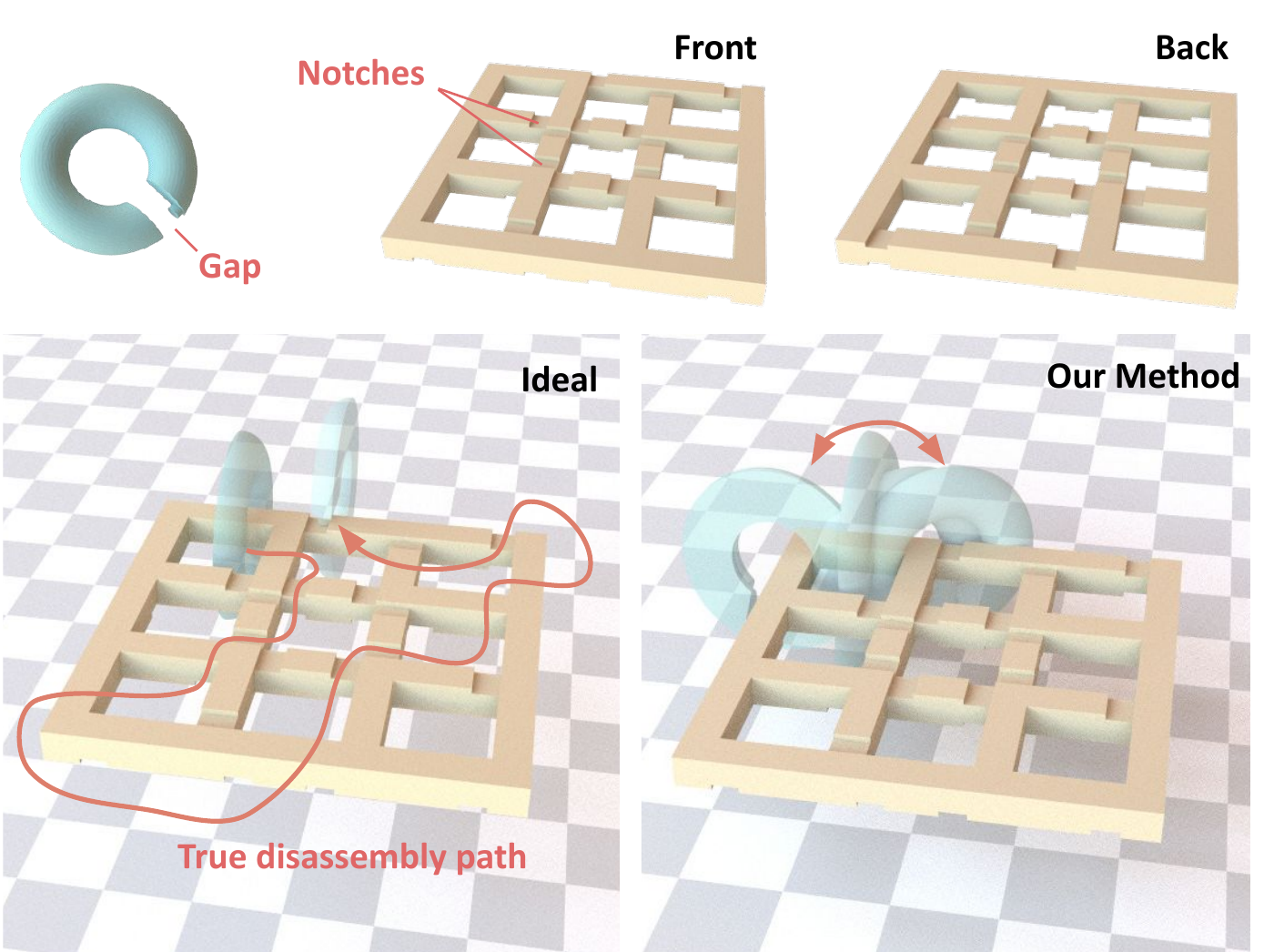}
    \caption{A failure case of our method disassembling a maze (geometries shown in the top row). The goal is to disassemble the ring by navigating through multiple narrow notches of the maze grid (bottom left). Our method finds difficulty in aligning the gap of the ring with notches of the grid, so the ring gets stuck in the initial local region (bottom right).}
    \label{fig:failure_maze}
    \vspace{-1em}
\end{figure}

\paragraph{Complex action sequences in path planning}
Assemblies that require very complex sequences of actions could be time-consuming for our BFS-guided method to explore. For example, in Figure~\ref{fig:failure_maze}, we test our method on a challenging maze-like assembly from \citet{zhang2020c}. Here, the motion space is less constrained but the solution space is extremely narrow. It requires complex action sequences to navigate the ring through multiple tunnels by correctly aligning the gap of the ring with the notches of the maze grid.

\begin{figure}
    \centering
    \includegraphics[width=0.45\textwidth]{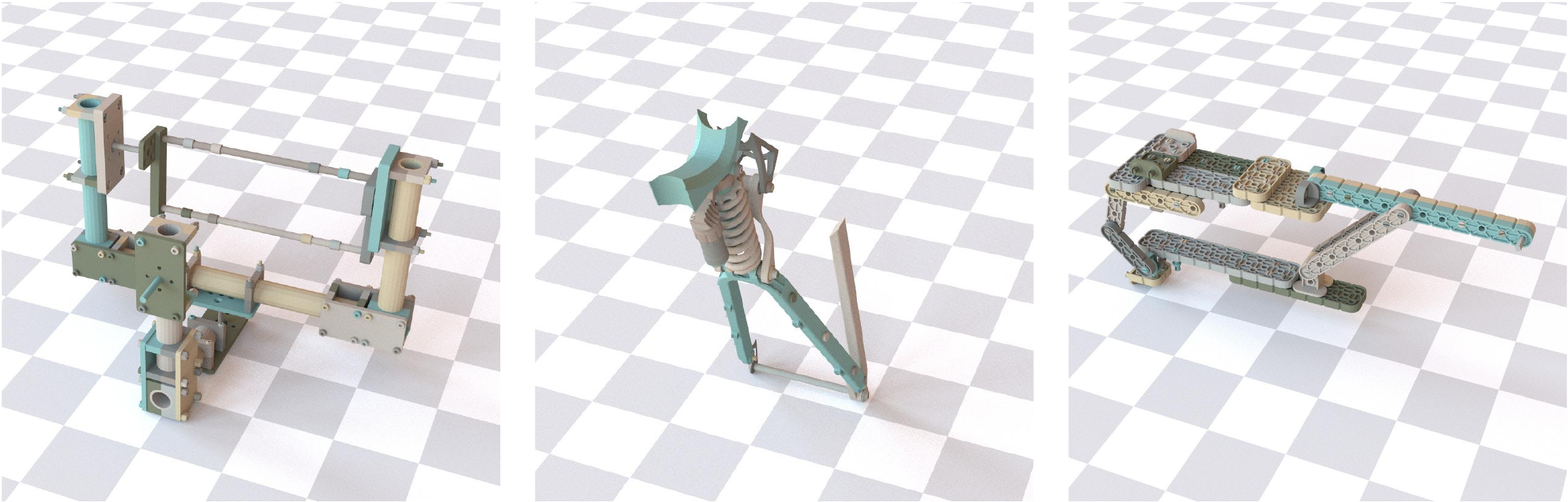}
    \caption{Large assemblies with several hundreds of parts and rich contacts where all tested methods reach the timeout of 2 hours.}
    \label{fig:failure_large}
    \vspace{-1em}
\end{figure}

\paragraph{Computational complexity on large assemblies}
Both our method and baseline methods fail on some large assemblies. This is due to larger assemblies having much richer contacts that typically result in a super-linear increase in the time cost of both the simulation and sequence planning. Figure~\ref{fig:failure_large} shows several examples where all methods timeout after 2 hours.

The sequence planning complexity is $O(M^2)$ for $M$-part sequential assemblies. However, for using sub-assemblies or interlocking assemblies, the complexity will grow to $O(M!)$ due to the combinatorial search for multiple parts to disassemble together in each step. Therefore, our naive extension is by no means an efficient method for large assemblies that involve sub-assemblies or interlock. In this case, heuristics or learning-based methods can potentially be adopted to speed up the sequence search.

\section{Conclusion and Future Work}

Planning for assemblies with arbitrary size, shape and motion is a long-standing and challenging problem. 
In this paper we have introduced a novel physics-based method for assembly planning. By providing a new method, custom simulation approach, and large-scale dataset we hope to enable future work that solves real world assembly and disassembly problems.
We envision several key extensions would make assembly planning more general and efficient.

First, our assembly-by-disassembly strategy limits the assemblies to being rigid only.
However, geometric-based approaches cannot handle deformable objects as well since they cannot model the physical deformation. Therefore, we believe it is interesting to explore further the direction of physics-based planning for generalization to deformable assemblies, such as the snap-fit assembly. 

Second, it is essential to leverage geometric information beyond physical feedback in assembly planning. In contrast to the exhaustive search adopted by our method and baseline methods, humans can instantly infer plausible disassembly sequences and motions through vision and avoid wasting effort trying blocked parts or moving parts towards dead ends.

Finally, to facilitate research in robotic assembly, one exciting future work is adding robotic arms in our simulation to manipulate assemblies following the planned path generated by our approach. We believe it is promising to extend the capability of performing complex assembly autonomously and flexibly on real robots.

\begin{acks}

We thank the anonymous reviewers for their helpful comments in revising the paper. We thank Yijiang Huang, Tao Chen, and Tao Du for their valuable feedback on research idea formulation and Timothy Ebinger for providing part of the assembly models in \citet{ebinger2018general}. This work was supported in part by the National Science Foundation (CAREER-1846368).

\end{acks}

\bibliographystyle{ACM-Reference-Format}
\bibliography{main}


\begin{thebibliography}{70}


\ifx \showCODEN    \undefined \def \showCODEN     #1{\unskip}     \fi
\ifx \showDOI      \undefined \def \showDOI       #1{#1}\fi
\ifx \showISBNx    \undefined \def \showISBNx     #1{\unskip}     \fi
\ifx \showISBNxiii \undefined \def \showISBNxiii  #1{\unskip}     \fi
\ifx \showISSN     \undefined \def \showISSN      #1{\unskip}     \fi
\ifx \showLCCN     \undefined \def \showLCCN      #1{\unskip}     \fi
\ifx \shownote     \undefined \def \shownote      #1{#1}          \fi
\ifx \showarticletitle \undefined \def \showarticletitle #1{#1}   \fi
\ifx \showURL      \undefined \def \showURL       {\relax}        \fi
\providecommand\bibfield[2]{#2}
\providecommand\bibinfo[2]{#2}
\providecommand\natexlab[1]{#1}
\providecommand\showeprint[2][]{arXiv:#2}

\bibitem[\protect\citeauthoryear{Agrawala, Phan, Heiser, Haymaker, Klingner,
  Hanrahan, and Tversky}{Agrawala et~al\mbox{.}}{2003}]%
        {agrawala2003designing}
\bibfield{author}{\bibinfo{person}{Maneesh Agrawala}, \bibinfo{person}{Doantam
  Phan}, \bibinfo{person}{Julie Heiser}, \bibinfo{person}{John Haymaker},
  \bibinfo{person}{Jeff Klingner}, \bibinfo{person}{Pat Hanrahan}, {and}
  \bibinfo{person}{Barbara Tversky}.} \bibinfo{year}{2003}\natexlab{}.
\newblock \showarticletitle{Designing effective step-by-step assembly
  instructions}.
\newblock \bibinfo{journal}{\emph{ACM Transactions on Graphics (TOG)}}
  \bibinfo{volume}{22}, \bibinfo{number}{3} (\bibinfo{year}{2003}),
  \bibinfo{pages}{828--837}.
\newblock


\bibitem[\protect\citeauthoryear{Aguinaga, Borro, and Matey}{Aguinaga
  et~al\mbox{.}}{2008}]%
        {aguinaga2008parallel}
\bibfield{author}{\bibinfo{person}{Iker Aguinaga}, \bibinfo{person}{Diego
  Borro}, {and} \bibinfo{person}{Luis Matey}.} \bibinfo{year}{2008}\natexlab{}.
\newblock \showarticletitle{Parallel RRT-based path planning for selective
  disassembly planning}.
\newblock \bibinfo{journal}{\emph{The International Journal of Advanced
  Manufacturing Technology}} \bibinfo{volume}{36}, \bibinfo{number}{11}
  (\bibinfo{year}{2008}), \bibinfo{pages}{1221--1233}.
\newblock


\bibitem[\protect\citeauthoryear{Aleotti and Caselli}{Aleotti and
  Caselli}{2009}]%
        {aleotti2009efficient}
\bibfield{author}{\bibinfo{person}{Jacopo Aleotti} {and}
  \bibinfo{person}{Stefano Caselli}.} \bibinfo{year}{2009}\natexlab{}.
\newblock \showarticletitle{Efficient planning of disassembly sequences in
  physics-based animation}. In \bibinfo{booktitle}{\emph{2009 IEEE/RSJ
  International Conference on Intelligent Robots and Systems}}. IEEE,
  \bibinfo{pages}{87--92}.
\newblock


\bibitem[\protect\citeauthoryear{Andrychowicz, Baker, Chociej, Józefowicz,
  McGrew, Pachocki, Petron, Plappert, Powell, Ray, Schneider, Sidor, Tobin,
  Welinder, Weng, and Zaremba}{Andrychowicz et~al\mbox{.}}{2020}]%
        {openai2020dexterous}
\bibfield{author}{\bibinfo{person}{OpenAI :~Marcin Andrychowicz},
  \bibinfo{person}{Bowen Baker}, \bibinfo{person}{Maciek Chociej},
  \bibinfo{person}{Rafal Józefowicz}, \bibinfo{person}{Bob McGrew},
  \bibinfo{person}{Jakub Pachocki}, \bibinfo{person}{Arthur Petron},
  \bibinfo{person}{Matthias Plappert}, \bibinfo{person}{Glenn Powell},
  \bibinfo{person}{Alex Ray}, \bibinfo{person}{Jonas Schneider},
  \bibinfo{person}{Szymon Sidor}, \bibinfo{person}{Josh Tobin},
  \bibinfo{person}{Peter Welinder}, \bibinfo{person}{Lilian Weng}, {and}
  \bibinfo{person}{Wojciech Zaremba}.} \bibinfo{year}{2020}\natexlab{}.
\newblock \showarticletitle{Learning dexterous in-hand manipulation}.
\newblock \bibinfo{journal}{\emph{The International Journal of Robotics
  Research}} \bibinfo{volume}{39}, \bibinfo{number}{1} (\bibinfo{year}{2020}),
  \bibinfo{pages}{3--20}.
\newblock


\bibitem[\protect\citeauthoryear{Brockman, Cheung, Pettersson, Schneider,
  Schulman, Tang, and Zaremba}{Brockman et~al\mbox{.}}{2016}]%
        {1606.01540}
\bibfield{author}{\bibinfo{person}{Greg Brockman}, \bibinfo{person}{Vicki
  Cheung}, \bibinfo{person}{Ludwig Pettersson}, \bibinfo{person}{Jonas
  Schneider}, \bibinfo{person}{John Schulman}, \bibinfo{person}{Jie Tang},
  {and} \bibinfo{person}{Wojciech Zaremba}.} \bibinfo{year}{2016}\natexlab{}.
\newblock \bibinfo{title}{OpenAI Gym}.
\newblock
\newblock
\showeprint{arXiv:1606.01540}


\bibitem[\protect\citeauthoryear{Chang, Funkhouser, Guibas, Hanrahan, Huang,
  Li, Savarese, Savva, Song, Su, et~al\mbox{.}}{Chang et~al\mbox{.}}{2015}]%
        {chang2015shapenet}
\bibfield{author}{\bibinfo{person}{Angel~X Chang}, \bibinfo{person}{Thomas
  Funkhouser}, \bibinfo{person}{Leonidas Guibas}, \bibinfo{person}{Pat
  Hanrahan}, \bibinfo{person}{Qixing Huang}, \bibinfo{person}{Zimo Li},
  \bibinfo{person}{Silvio Savarese}, \bibinfo{person}{Manolis Savva},
  \bibinfo{person}{Shuran Song}, \bibinfo{person}{Hao Su}, {et~al\mbox{.}}}
  \bibinfo{year}{2015}\natexlab{}.
\newblock \showarticletitle{Shapenet: An information-rich 3d model repository}.
\newblock \bibinfo{journal}{\emph{arXiv:1512.03012}} (\bibinfo{year}{2015}).
\newblock


\bibitem[\protect\citeauthoryear{Chebotar, Handa, Makoviychuk, Macklin, Issac,
  Ratliff, and Fox}{Chebotar et~al\mbox{.}}{2019}]%
        {chebotar2019closing}
\bibfield{author}{\bibinfo{person}{Yevgen Chebotar}, \bibinfo{person}{Ankur
  Handa}, \bibinfo{person}{Viktor Makoviychuk}, \bibinfo{person}{Miles
  Macklin}, \bibinfo{person}{Jan Issac}, \bibinfo{person}{Nathan Ratliff},
  {and} \bibinfo{person}{Dieter Fox}.} \bibinfo{year}{2019}\natexlab{}.
\newblock \showarticletitle{Closing the sim-to-real loop: Adapting simulation
  randomization with real world experience}. In \bibinfo{booktitle}{\emph{2019
  International Conference on Robotics and Automation (ICRA)}}. IEEE,
  \bibinfo{pages}{8973--8979}.
\newblock


\bibitem[\protect\citeauthoryear{Chen, Zhou, Cai, Lai, Lin, and Menassa}{Chen
  et~al\mbox{.}}{2006}]%
        {chen2006framework}
\bibfield{author}{\bibinfo{person}{Guanlong Chen}, \bibinfo{person}{Jiangqi
  Zhou}, \bibinfo{person}{Wayne Cai}, \bibinfo{person}{Xinmin Lai},
  \bibinfo{person}{Zhongqin Lin}, {and} \bibinfo{person}{Roland Menassa}.}
  \bibinfo{year}{2006}\natexlab{}.
\newblock \showarticletitle{A framework for an automotive body assembly process
  design system}.
\newblock \bibinfo{journal}{\emph{Computer-Aided Design}} \bibinfo{volume}{38},
  \bibinfo{number}{5} (\bibinfo{year}{2006}), \bibinfo{pages}{531--539}.
\newblock


\bibitem[\protect\citeauthoryear{Chen, Wan, and Harada}{Chen
  et~al\mbox{.}}{2021a}]%
        {chen2021planning}
\bibfield{author}{\bibinfo{person}{Hao Chen}, \bibinfo{person}{Weiwei Wan},
  {and} \bibinfo{person}{Kensuke Harada}.} \bibinfo{year}{2021}\natexlab{a}.
\newblock \showarticletitle{Planning to Build Soma Blocks Using a Dual-arm
  Robot}. In \bibinfo{booktitle}{\emph{2021 IEEE International Conference on
  Development and Learning (ICDL)}}. IEEE, \bibinfo{pages}{1--7}.
\newblock


\bibitem[\protect\citeauthoryear{Chen, Xu, and Agrawal}{Chen
  et~al\mbox{.}}{2021b}]%
        {chen2021simple}
\bibfield{author}{\bibinfo{person}{Tao Chen}, \bibinfo{person}{Jie Xu}, {and}
  \bibinfo{person}{Pulkit Agrawal}.} \bibinfo{year}{2021}\natexlab{b}.
\newblock \showarticletitle{A Simple Method for Complex In-hand Manipulation}.
\newblock \bibinfo{journal}{\emph{Conference on Robot Learning}}
  (\bibinfo{year}{2021}).
\newblock


\bibitem[\protect\citeauthoryear{Chen, Tai, Deng, and Hsieh}{Chen
  et~al\mbox{.}}{2008}]%
        {chen2008three}
\bibfield{author}{\bibinfo{person}{Wen-Chin Chen}, \bibinfo{person}{Pei-Hao
  Tai}, \bibinfo{person}{Wei-Jaw Deng}, {and} \bibinfo{person}{Ling-Feng
  Hsieh}.} \bibinfo{year}{2008}\natexlab{}.
\newblock \showarticletitle{A three-stage integrated approach for assembly
  sequence planning using neural networks}.
\newblock \bibinfo{journal}{\emph{Expert Systems with Applications}}
  \bibinfo{volume}{34}, \bibinfo{number}{3} (\bibinfo{year}{2008}),
  \bibinfo{pages}{1777--1786}.
\newblock


\bibitem[\protect\citeauthoryear{Coulom}{Coulom}{2006}]%
        {coulom2006efficient}
\bibfield{author}{\bibinfo{person}{R{\'e}mi Coulom}.}
  \bibinfo{year}{2006}\natexlab{}.
\newblock \showarticletitle{Efficient selectivity and backup operators in
  Monte-Carlo tree search}. In \bibinfo{booktitle}{\emph{International
  conference on computers and games}}. Springer, \bibinfo{pages}{72--83}.
\newblock


\bibitem[\protect\citeauthoryear{Coumans and Bai}{Coumans and Bai}{2016}]%
        {coumans2016pybullet}
\bibfield{author}{\bibinfo{person}{Erwin Coumans} {and} \bibinfo{person}{Yunfei
  Bai}.} \bibinfo{year}{2016}\natexlab{}.
\newblock \showarticletitle{Pybullet, a python module for physics simulation
  for games, robotics and machine learning}.
\newblock  (\bibinfo{year}{2016}).
\newblock


\bibitem[\protect\citeauthoryear{De~Mello and Sanderson}{De~Mello and
  Sanderson}{1990}]%
        {de1990and}
\bibfield{author}{\bibinfo{person}{LS~Homem De~Mello} {and}
  \bibinfo{person}{Arthur~C Sanderson}.} \bibinfo{year}{1990}\natexlab{}.
\newblock \showarticletitle{AND/OR graph representation of assembly plans}.
\newblock \bibinfo{journal}{\emph{IEEE Transactions on robotics and
  automation}} \bibinfo{volume}{6}, \bibinfo{number}{2} (\bibinfo{year}{1990}),
  \bibinfo{pages}{188--199}.
\newblock


\bibitem[\protect\citeauthoryear{De~Winter, EI~Makrini, Van~de Perre, Now{\'e},
  Verstraten, and Vanderborght}{De~Winter et~al\mbox{.}}{2021}]%
        {de2021autonomous}
\bibfield{author}{\bibinfo{person}{Joris De~Winter}, \bibinfo{person}{Ilias
  EI~Makrini}, \bibinfo{person}{Greet Van~de Perre}, \bibinfo{person}{Ann
  Now{\'e}}, \bibinfo{person}{Tom Verstraten}, {and} \bibinfo{person}{Bram
  Vanderborght}.} \bibinfo{year}{2021}\natexlab{}.
\newblock \showarticletitle{Autonomous assembly planning of demonstrated skills
  with reinforcement learning in simulation}.
\newblock \bibinfo{journal}{\emph{Autonomous Robots}} \bibinfo{volume}{45},
  \bibinfo{number}{8} (\bibinfo{year}{2021}), \bibinfo{pages}{1097--1110}.
\newblock


\bibitem[\protect\citeauthoryear{Ebinger, Kaden, Thomas, Andre, Amato, and
  Thomas}{Ebinger et~al\mbox{.}}{2018}]%
        {ebinger2018general}
\bibfield{author}{\bibinfo{person}{Timothy Ebinger}, \bibinfo{person}{Sascha
  Kaden}, \bibinfo{person}{Shawna Thomas}, \bibinfo{person}{Robert Andre},
  \bibinfo{person}{Nancy~M Amato}, {and} \bibinfo{person}{Ulrike Thomas}.}
  \bibinfo{year}{2018}\natexlab{}.
\newblock \showarticletitle{A general and flexible search framework for
  disassembly planning}. In \bibinfo{booktitle}{\emph{2018 IEEE International
  Conference on Robotics and Automation (ICRA)}}. IEEE,
  \bibinfo{pages}{3548--3555}.
\newblock


\bibitem[\protect\citeauthoryear{Fan, Luo, and Tomizuka}{Fan
  et~al\mbox{.}}{2019}]%
        {fan2019learning}
\bibfield{author}{\bibinfo{person}{Yongxiang Fan}, \bibinfo{person}{Jieliang
  Luo}, {and} \bibinfo{person}{Masayoshi Tomizuka}.}
  \bibinfo{year}{2019}\natexlab{}.
\newblock \showarticletitle{A learning framework for high precision industrial
  assembly}. In \bibinfo{booktitle}{\emph{2019 International Conference on
  Robotics and Automation (ICRA)}}. IEEE, \bibinfo{pages}{811--817}.
\newblock


\bibitem[\protect\citeauthoryear{Geilinger, Hahn, Zehnder, B\"{a}cher,
  Thomaszewski, and Coros}{Geilinger et~al\mbox{.}}{2020}]%
        {geilinger2020add}
\bibfield{author}{\bibinfo{person}{Moritz Geilinger}, \bibinfo{person}{David
  Hahn}, \bibinfo{person}{Jonas Zehnder}, \bibinfo{person}{Moritz B\"{a}cher},
  \bibinfo{person}{Bernhard Thomaszewski}, {and} \bibinfo{person}{Stelian
  Coros}.} \bibinfo{year}{2020}\natexlab{}.
\newblock \showarticletitle{ADD: Analytically Differentiable Dynamics for
  Multi-Body Systems with Frictional Contact}.
\newblock \bibinfo{journal}{\emph{ACM Trans. Graph.}} \bibinfo{volume}{39},
  \bibinfo{number}{6}, Article \bibinfo{articleno}{190} (\bibinfo{date}{Nov.}
  \bibinfo{year}{2020}), \bibinfo{numpages}{15}~pages.
\newblock


\bibitem[\protect\citeauthoryear{Ghandi and Masehian}{Ghandi and
  Masehian}{2015}]%
        {ghandi2015review}
\bibfield{author}{\bibinfo{person}{Somaye Ghandi} {and} \bibinfo{person}{Ellips
  Masehian}.} \bibinfo{year}{2015}\natexlab{}.
\newblock \showarticletitle{Review and taxonomies of assembly and disassembly
  path planning problems and approaches}.
\newblock \bibinfo{journal}{\emph{Computer-Aided Design}}  \bibinfo{volume}{67}
  (\bibinfo{year}{2015}), \bibinfo{pages}{58--86}.
\newblock


\bibitem[\protect\citeauthoryear{Halperin, Latombe, and Wilson}{Halperin
  et~al\mbox{.}}{2000}]%
        {halperin2000general}
\bibfield{author}{\bibinfo{person}{Dan Halperin}, \bibinfo{person}{J-C
  Latombe}, {and} \bibinfo{person}{Randall~H Wilson}.}
  \bibinfo{year}{2000}\natexlab{}.
\newblock \showarticletitle{A general framework for assembly planning: The
  motion space approach}.
\newblock \bibinfo{journal}{\emph{Algorithmica}} \bibinfo{volume}{26},
  \bibinfo{number}{3} (\bibinfo{year}{2000}), \bibinfo{pages}{577--601}.
\newblock


\bibitem[\protect\citeauthoryear{Hartmann, Orthey, Driess, Oguz, and
  Toussaint}{Hartmann et~al\mbox{.}}{2021}]%
        {hartmann2021long}
\bibfield{author}{\bibinfo{person}{Valentin~Noah Hartmann},
  \bibinfo{person}{Andreas Orthey}, \bibinfo{person}{Danny Driess},
  \bibinfo{person}{Ozgur~S Oguz}, {and} \bibinfo{person}{Marc Toussaint}.}
  \bibinfo{year}{2021}\natexlab{}.
\newblock \showarticletitle{Long-horizon multi-robot rearrangement planning for
  construction assembly}.
\newblock \bibinfo{journal}{\emph{arXiv preprint arXiv:2106.02489}}
  (\bibinfo{year}{2021}).
\newblock


\bibitem[\protect\citeauthoryear{Homem~de Mello and Sanderson}{Homem~de Mello
  and Sanderson}{1991}]%
        {demello1991}
\bibfield{author}{\bibinfo{person}{L.S. Homem~de Mello} {and}
  \bibinfo{person}{A.C. Sanderson}.} \bibinfo{year}{1991}\natexlab{}.
\newblock \showarticletitle{A correct and complete algorithm for the generation
  of mechanical assembly sequences}.
\newblock \bibinfo{journal}{\emph{IEEE Transactions on Robotics and
  Automation}} \bibinfo{volume}{7}, \bibinfo{number}{2} (\bibinfo{year}{1991}),
  \bibinfo{pages}{228--240}.
\newblock
\urldef\tempurl%
\url{https://doi.org/10.1109/70.75905}
\showDOI{\tempurl}


\bibitem[\protect\citeauthoryear{Hou, Fei, Deng, and Xu}{Hou
  et~al\mbox{.}}{2020}]%
        {hou2020data}
\bibfield{author}{\bibinfo{person}{Zhimin Hou}, \bibinfo{person}{Jiajun Fei},
  \bibinfo{person}{Yuelin Deng}, {and} \bibinfo{person}{Jing Xu}.}
  \bibinfo{year}{2020}\natexlab{}.
\newblock \showarticletitle{Data-efficient hierarchical reinforcement learning
  for robotic assembly control applications}.
\newblock \bibinfo{journal}{\emph{IEEE Transactions on Industrial Electronics}}
  \bibinfo{volume}{68}, \bibinfo{number}{11} (\bibinfo{year}{2020}),
  \bibinfo{pages}{11565--11575}.
\newblock


\bibitem[\protect\citeauthoryear{Hsu, Latombe, Motwani, and Kavraki}{Hsu
  et~al\mbox{.}}{1999}]%
        {hsu1999capturing}
\bibfield{author}{\bibinfo{person}{David Hsu}, \bibinfo{person}{Jean-Claude
  Latombe}, \bibinfo{person}{Rajeev Motwani}, {and} \bibinfo{person}{Lydia~E
  Kavraki}.} \bibinfo{year}{1999}\natexlab{}.
\newblock \showarticletitle{Capturing the connectivity of high-dimensional
  geometric spaces by parallelizable random sampling techniques}.
\newblock In \bibinfo{booktitle}{\emph{Advances in randomized parallel
  computing}}. \bibinfo{publisher}{Springer}, \bibinfo{pages}{159--182}.
\newblock


\bibitem[\protect\citeauthoryear{Hu, Li, Van~Kaick, Shamir, Zhang, and
  Huang}{Hu et~al\mbox{.}}{2017}]%
        {hu2017learning}
\bibfield{author}{\bibinfo{person}{Ruizhen Hu}, \bibinfo{person}{Wenchao Li},
  \bibinfo{person}{Oliver Van~Kaick}, \bibinfo{person}{Ariel Shamir},
  \bibinfo{person}{Hao Zhang}, {and} \bibinfo{person}{Hui Huang}.}
  \bibinfo{year}{2017}\natexlab{}.
\newblock \showarticletitle{Learning to predict part mobility from a single
  static snapshot}.
\newblock \bibinfo{journal}{\emph{ACM Transactions on Graphics (TOG)}}
  \bibinfo{volume}{36}, \bibinfo{number}{6} (\bibinfo{year}{2017}),
  \bibinfo{pages}{1--13}.
\newblock


\bibitem[\protect\citeauthoryear{Huang, Garrett, Ting, Parascho, and
  Mueller}{Huang et~al\mbox{.}}{2021}]%
        {huang2021robotic}
\bibfield{author}{\bibinfo{person}{Yijiang Huang}, \bibinfo{person}{Caelan~R
  Garrett}, \bibinfo{person}{Ian Ting}, \bibinfo{person}{Stefana Parascho},
  {and} \bibinfo{person}{Caitlin~T Mueller}.} \bibinfo{year}{2021}\natexlab{}.
\newblock \showarticletitle{Robotic additive construction of bar structures:
  Unified sequence and motion planning}.
\newblock \bibinfo{journal}{\emph{Construction Robotics}} \bibinfo{volume}{5},
  \bibinfo{number}{2} (\bibinfo{year}{2021}), \bibinfo{pages}{115--130}.
\newblock


\bibitem[\protect\citeauthoryear{Jones, Hildreth, Chen, Baran, Kim, and
  Schulz}{Jones et~al\mbox{.}}{2021}]%
        {jones2021automate}
\bibfield{author}{\bibinfo{person}{Benjamin Jones}, \bibinfo{person}{Dalton
  Hildreth}, \bibinfo{person}{Duowen Chen}, \bibinfo{person}{Ilya Baran},
  \bibinfo{person}{Vladimir~G. Kim}, {and} \bibinfo{person}{Adriana Schulz}.}
  \bibinfo{year}{2021}\natexlab{}.
\newblock \showarticletitle{AutoMate: A Dataset and Learning Approach for
  Automatic Mating of CAD Assemblies}.
\newblock \bibinfo{journal}{\emph{ACM Transactions on Graphics (TOG)}}
  \bibinfo{volume}{40}, \bibinfo{number}{6} (\bibinfo{year}{2021}).
\newblock


\bibitem[\protect\citeauthoryear{Kavraki, Svestka, Latombe, and
  Overmars}{Kavraki et~al\mbox{.}}{1996}]%
        {kavraki1996probabilistic}
\bibfield{author}{\bibinfo{person}{Lydia~E Kavraki}, \bibinfo{person}{Petr
  Svestka}, \bibinfo{person}{J-C Latombe}, {and} \bibinfo{person}{Mark~H
  Overmars}.} \bibinfo{year}{1996}\natexlab{}.
\newblock \showarticletitle{Probabilistic roadmaps for path planning in
  high-dimensional configuration spaces}.
\newblock \bibinfo{journal}{\emph{IEEE transactions on Robotics and
  Automation}} \bibinfo{volume}{12}, \bibinfo{number}{4}
  (\bibinfo{year}{1996}), \bibinfo{pages}{566--580}.
\newblock


\bibitem[\protect\citeauthoryear{Koch, Matveev, Jiang, Williams, Artemov,
  Burnaev, Alexa, Zorin, and Panozzo}{Koch et~al\mbox{.}}{2019}]%
        {koch2019abc}
\bibfield{author}{\bibinfo{person}{Sebastian Koch}, \bibinfo{person}{Albert
  Matveev}, \bibinfo{person}{Zhongshi Jiang}, \bibinfo{person}{Francis
  Williams}, \bibinfo{person}{Alexey Artemov}, \bibinfo{person}{Evgeny
  Burnaev}, \bibinfo{person}{Marc Alexa}, \bibinfo{person}{Denis Zorin}, {and}
  \bibinfo{person}{Daniele Panozzo}.} \bibinfo{year}{2019}\natexlab{}.
\newblock \showarticletitle{ABC: A big CAD model dataset for geometric deep
  learning}. In \bibinfo{booktitle}{\emph{IEEE Conference on Computer Vision
  and Pattern Recognition (CVPR)}}. \bibinfo{pages}{9601--9611}.
\newblock


\bibitem[\protect\citeauthoryear{Kuffner and LaValle}{Kuffner and
  LaValle}{2000}]%
        {kuffner2000rrt}
\bibfield{author}{\bibinfo{person}{James~J Kuffner} {and}
  \bibinfo{person}{Steven~M LaValle}.} \bibinfo{year}{2000}\natexlab{}.
\newblock \showarticletitle{RRT-connect: An efficient approach to single-query
  path planning}. In \bibinfo{booktitle}{\emph{Proceedings 2000 ICRA.
  Millennium Conference. IEEE International Conference on Robotics and
  Automation. Symposia Proceedings (Cat. No. 00CH37065)}},
  Vol.~\bibinfo{volume}{2}. IEEE, \bibinfo{pages}{995--1001}.
\newblock


\bibitem[\protect\citeauthoryear{LaValle et~al\mbox{.}}{LaValle
  et~al\mbox{.}}{1998}]%
        {lavalle1998rapidly}
\bibfield{author}{\bibinfo{person}{Steven~M LaValle} {et~al\mbox{.}}}
  \bibinfo{year}{1998}\natexlab{}.
\newblock \showarticletitle{Rapidly-exploring random trees: A new tool for path
  planning}.
\newblock  (\bibinfo{year}{1998}).
\newblock


\bibitem[\protect\citeauthoryear{Le, Cort{\'e}s, and Sim{\'e}on}{Le
  et~al\mbox{.}}{2009}]%
        {le2009path}
\bibfield{author}{\bibinfo{person}{Duc~Thanh Le}, \bibinfo{person}{Juan
  Cort{\'e}s}, {and} \bibinfo{person}{Thierry Sim{\'e}on}.}
  \bibinfo{year}{2009}\natexlab{}.
\newblock \showarticletitle{A path planning approach to (dis) assembly
  sequencing}. In \bibinfo{booktitle}{\emph{2009 IEEE International Conference
  on Automation Science and Engineering}}. IEEE, \bibinfo{pages}{286--291}.
\newblock


\bibitem[\protect\citeauthoryear{Luo and Li}{Luo and Li}{2021}]%
        {luo2021learning}
\bibfield{author}{\bibinfo{person}{Jieliang Luo} {and} \bibinfo{person}{Hui
  Li}.} \bibinfo{year}{2021}\natexlab{}.
\newblock \showarticletitle{A Learning Approach to Robot-Agnostic Force-Guided
  High Precision Assembly}. In \bibinfo{booktitle}{\emph{2021 IEEE/RSJ
  International Conference on Intelligent Robots and Systems (IROS)}}. IEEE,
  \bibinfo{pages}{2151--2157}.
\newblock


\bibitem[\protect\citeauthoryear{Makoviychuk, Wawrzyniak, Guo, Lu, Storey,
  Macklin, Hoeller, Rudin, Allshire, Handa, et~al\mbox{.}}{Makoviychuk
  et~al\mbox{.}}{2021}]%
        {makoviychuk2021isaac}
\bibfield{author}{\bibinfo{person}{Viktor Makoviychuk}, \bibinfo{person}{Lukasz
  Wawrzyniak}, \bibinfo{person}{Yunrong Guo}, \bibinfo{person}{Michelle Lu},
  \bibinfo{person}{Kier Storey}, \bibinfo{person}{Miles Macklin},
  \bibinfo{person}{David Hoeller}, \bibinfo{person}{Nikita Rudin},
  \bibinfo{person}{Arthur Allshire}, \bibinfo{person}{Ankur Handa},
  {et~al\mbox{.}}} \bibinfo{year}{2021}\natexlab{}.
\newblock \showarticletitle{Isaac Gym: High Performance GPU Based Physics
  Simulation For Robot Learning}. In \bibinfo{booktitle}{\emph{Thirty-fifth
  Conference on Neural Information Processing Systems Datasets and Benchmarks
  Track (Round 2)}}.
\newblock


\bibitem[\protect\citeauthoryear{Masehian and Ghandi}{Masehian and
  Ghandi}{2021}]%
        {masehian2021assembly}
\bibfield{author}{\bibinfo{person}{Ellips Masehian} {and}
  \bibinfo{person}{Somay{\'e} Ghandi}.} \bibinfo{year}{2021}\natexlab{}.
\newblock \showarticletitle{Assembly sequence and path planning for monotone
  and nonmonotone assemblies with rigid and flexible parts}.
\newblock \bibinfo{journal}{\emph{Robotics and Computer-Integrated
  Manufacturing}}  \bibinfo{volume}{72} (\bibinfo{year}{2021}),
  \bibinfo{pages}{102180}.
\newblock


\bibitem[\protect\citeauthoryear{Melckenbeeck, Burggraeve, Van~Doninck,
  Vancraen, and Rosich}{Melckenbeeck et~al\mbox{.}}{2020}]%
        {melckenbeeck2020optimal}
\bibfield{author}{\bibinfo{person}{Ine Melckenbeeck}, \bibinfo{person}{Sofie
  Burggraeve}, \bibinfo{person}{Bart Van~Doninck}, \bibinfo{person}{Jeroen
  Vancraen}, {and} \bibinfo{person}{Albert Rosich}.}
  \bibinfo{year}{2020}\natexlab{}.
\newblock \showarticletitle{Optimal assembly sequence based on design for
  assembly (DFA) rules}.
\newblock \bibinfo{journal}{\emph{Procedia CIRP}}  \bibinfo{volume}{91}
  (\bibinfo{year}{2020}), \bibinfo{pages}{646--652}.
\newblock


\bibitem[\protect\citeauthoryear{Mo, Zhu, Chang, Yi, Tripathi, Guibas, and
  Su}{Mo et~al\mbox{.}}{2019}]%
        {mo2019partnet}
\bibfield{author}{\bibinfo{person}{Kaichun Mo}, \bibinfo{person}{Shilin Zhu},
  \bibinfo{person}{Angel~X Chang}, \bibinfo{person}{Li Yi},
  \bibinfo{person}{Subarna Tripathi}, \bibinfo{person}{Leonidas~J Guibas},
  {and} \bibinfo{person}{Hao Su}.} \bibinfo{year}{2019}\natexlab{}.
\newblock \showarticletitle{Partnet: A large-scale benchmark for fine-grained
  and hierarchical part-level 3d object understanding}. In
  \bibinfo{booktitle}{\emph{IEEE Conference on Computer Vision and Pattern
  Recognition (CVPR)}}. \bibinfo{pages}{909--918}.
\newblock


\bibitem[\protect\citeauthoryear{Moll, Kavraki, Rosell, et~al\mbox{.}}{Moll
  et~al\mbox{.}}{2017}]%
        {moll2017randomized}
\bibfield{author}{\bibinfo{person}{Mark Moll}, \bibinfo{person}{Lydia Kavraki},
  \bibinfo{person}{Jan Rosell}, {et~al\mbox{.}}}
  \bibinfo{year}{2017}\natexlab{}.
\newblock \showarticletitle{Randomized physics-based motion planning for
  grasping in cluttered and uncertain environments}.
\newblock \bibinfo{journal}{\emph{IEEE Robotics and Automation Letters}}
  \bibinfo{volume}{3}, \bibinfo{number}{2} (\bibinfo{year}{2017}),
  \bibinfo{pages}{712--719}.
\newblock


\bibitem[\protect\citeauthoryear{Narang, Storey, Akinola, Macklin, Reist,
  Wawrzyniak, Guo, Moravanszky, State, Lu, et~al\mbox{.}}{Narang
  et~al\mbox{.}}{2022}]%
        {narang2022factory}
\bibfield{author}{\bibinfo{person}{Yashraj Narang}, \bibinfo{person}{Kier
  Storey}, \bibinfo{person}{Iretiayo Akinola}, \bibinfo{person}{Miles Macklin},
  \bibinfo{person}{Philipp Reist}, \bibinfo{person}{Lukasz Wawrzyniak},
  \bibinfo{person}{Yunrong Guo}, \bibinfo{person}{Adam Moravanszky},
  \bibinfo{person}{Gavriel State}, \bibinfo{person}{Michelle Lu},
  {et~al\mbox{.}}} \bibinfo{year}{2022}\natexlab{}.
\newblock \showarticletitle{Factory: Fast Contact for Robotic Assembly}.
\newblock \bibinfo{journal}{\emph{arXiv preprint arXiv:2205.03532}}
  (\bibinfo{year}{2022}).
\newblock


\bibitem[\protect\citeauthoryear{Niu, Ding, and Xiong}{Niu
  et~al\mbox{.}}{2003}]%
        {niu2003hierarchical}
\bibfield{author}{\bibinfo{person}{Xinwen Niu}, \bibinfo{person}{Han Ding},
  {and} \bibinfo{person}{Youlun Xiong}.} \bibinfo{year}{2003}\natexlab{}.
\newblock \showarticletitle{A hierarchical approach to generating precedence
  graphs for assembly planning}.
\newblock \bibinfo{journal}{\emph{International Journal of Machine Tools and
  Manufacture}} \bibinfo{volume}{43}, \bibinfo{number}{14}
  (\bibinfo{year}{2003}), \bibinfo{pages}{1473--1486}.
\newblock


\bibitem[\protect\citeauthoryear{Ong, Chang, and Nee}{Ong
  et~al\mbox{.}}{2021}]%
        {ong2021product}
\bibfield{author}{\bibinfo{person}{SK Ong}, \bibinfo{person}{MML Chang}, {and}
  \bibinfo{person}{AYC Nee}.} \bibinfo{year}{2021}\natexlab{}.
\newblock \showarticletitle{Product disassembly sequence planning:
  state-of-the-art, challenges, opportunities and future directions}.
\newblock \bibinfo{journal}{\emph{International Journal of Production
  Research}} \bibinfo{volume}{59}, \bibinfo{number}{11} (\bibinfo{year}{2021}),
  \bibinfo{pages}{3493--3508}.
\newblock


\bibitem[\protect\citeauthoryear{Qin and Xu}{Qin and Xu}{2007}]%
        {qin2007assembly}
\bibfield{author}{\bibinfo{person}{YF Qin} {and} \bibinfo{person}{ZG Xu}.}
  \bibinfo{year}{2007}\natexlab{}.
\newblock \showarticletitle{Assembly process planning using a multi-objective
  optimization method}. In \bibinfo{booktitle}{\emph{Proceedings of the 2007
  IEEE international conference on mechatronics and automation, ICMA}},
  Vol.~\bibinfo{volume}{4303610}. \bibinfo{pages}{593--598}.
\newblock


\bibitem[\protect\citeauthoryear{Ramos, Rocha, and Vale}{Ramos
  et~al\mbox{.}}{1998}]%
        {ramos1998complexity}
\bibfield{author}{\bibinfo{person}{Carlos Ramos}, \bibinfo{person}{Joao Rocha},
  {and} \bibinfo{person}{Zita Vale}.} \bibinfo{year}{1998}\natexlab{}.
\newblock \showarticletitle{On the complexity of precedence graphs for assembly
  and task planning}.
\newblock \bibinfo{journal}{\emph{Computers in industry}} \bibinfo{volume}{36},
  \bibinfo{number}{1-2} (\bibinfo{year}{1998}), \bibinfo{pages}{101--111}.
\newblock


\bibitem[\protect\citeauthoryear{Rashid, Hutabarat, and Tiwari}{Rashid
  et~al\mbox{.}}{2012}]%
        {rashid2012review}
\bibfield{author}{\bibinfo{person}{Mohd Fadzil~Faisae Rashid},
  \bibinfo{person}{Windo Hutabarat}, {and} \bibinfo{person}{Ashutosh Tiwari}.}
  \bibinfo{year}{2012}\natexlab{}.
\newblock \showarticletitle{A review on assembly sequence planning and assembly
  line balancing optimisation using soft computing approaches}.
\newblock \bibinfo{journal}{\emph{The International Journal of Advanced
  Manufacturing Technology}} \bibinfo{volume}{59}, \bibinfo{number}{1}
  (\bibinfo{year}{2012}), \bibinfo{pages}{335--349}.
\newblock


\bibitem[\protect\citeauthoryear{Rold{\'a}n, Crespo, Mart{\'\i}n-Barrio,
  Pe{\~n}a-Tapia, and Barrientos}{Rold{\'a}n et~al\mbox{.}}{2019}]%
        {roldan2019training}
\bibfield{author}{\bibinfo{person}{Juan~Jes{\'u}s Rold{\'a}n},
  \bibinfo{person}{Elena Crespo}, \bibinfo{person}{Andr{\'e}s
  Mart{\'\i}n-Barrio}, \bibinfo{person}{Elena Pe{\~n}a-Tapia}, {and}
  \bibinfo{person}{Antonio Barrientos}.} \bibinfo{year}{2019}\natexlab{}.
\newblock \showarticletitle{A training system for Industry 4.0 operators in
  complex assemblies based on virtual reality and process mining}.
\newblock \bibinfo{journal}{\emph{Robotics and computer-integrated
  manufacturing}}  \bibinfo{volume}{59} (\bibinfo{year}{2019}),
  \bibinfo{pages}{305--316}.
\newblock


\bibitem[\protect\citeauthoryear{Santochi, Dini, and Failli}{Santochi
  et~al\mbox{.}}{2002}]%
        {santochi2002computer}
\bibfield{author}{\bibinfo{person}{Marco Santochi}, \bibinfo{person}{Gino
  Dini}, {and} \bibinfo{person}{Franco Failli}.}
  \bibinfo{year}{2002}\natexlab{}.
\newblock \showarticletitle{Computer aided disassembly planning: state of the
  art and perspectives}.
\newblock \bibinfo{journal}{\emph{CIRP Annals}} \bibinfo{volume}{51},
  \bibinfo{number}{2} (\bibinfo{year}{2002}), \bibinfo{pages}{507--529}.
\newblock


\bibitem[\protect\citeauthoryear{Sinano{\u{g}}lu and
  B{\"o}rkl{\"u}}{Sinano{\u{g}}lu and B{\"o}rkl{\"u}}{2005}]%
        {sinanouglu2005assembly}
\bibfield{author}{\bibinfo{person}{Cem Sinano{\u{g}}lu} {and}
  \bibinfo{person}{H~R{\i}za B{\"o}rkl{\"u}}.} \bibinfo{year}{2005}\natexlab{}.
\newblock \showarticletitle{An assembly sequence-planning system for mechanical
  parts using neural network}.
\newblock \bibinfo{journal}{\emph{Assembly Automation}} (\bibinfo{year}{2005}).
\newblock


\bibitem[\protect\citeauthoryear{Su}{Su}{2009}]%
        {su2009hierarchical}
\bibfield{author}{\bibinfo{person}{Qiang Su}.} \bibinfo{year}{2009}\natexlab{}.
\newblock \showarticletitle{A hierarchical approach on assembly sequence
  planning and optimal sequences analyzing}.
\newblock \bibinfo{journal}{\emph{Robotics and Computer-Integrated
  Manufacturing}} \bibinfo{volume}{25}, \bibinfo{number}{1}
  (\bibinfo{year}{2009}), \bibinfo{pages}{224--234}.
\newblock


\bibitem[\protect\citeauthoryear{Sucan and Kavraki}{Sucan and Kavraki}{2011}]%
        {sucan2011sampling}
\bibfield{author}{\bibinfo{person}{Ioan~A Sucan} {and} \bibinfo{person}{Lydia~E
  Kavraki}.} \bibinfo{year}{2011}\natexlab{}.
\newblock \showarticletitle{A sampling-based tree planner for systems with
  complex dynamics}.
\newblock \bibinfo{journal}{\emph{IEEE Transactions on Robotics}}
  \bibinfo{volume}{28}, \bibinfo{number}{1} (\bibinfo{year}{2011}),
  \bibinfo{pages}{116--131}.
\newblock


\bibitem[\protect\citeauthoryear{Sundaram, Remmler, and Amato}{Sundaram
  et~al\mbox{.}}{2001}]%
        {sundaram2001disassembly}
\bibfield{author}{\bibinfo{person}{Sujay Sundaram}, \bibinfo{person}{Ian
  Remmler}, {and} \bibinfo{person}{Nancy~M Amato}.}
  \bibinfo{year}{2001}\natexlab{}.
\newblock \showarticletitle{Disassembly sequencing using a motion planning
  approach}. In \bibinfo{booktitle}{\emph{Proceedings 2001 ICRA. IEEE
  International Conference on Robotics and Automation (Cat. No. 01CH37164)}},
  Vol.~\bibinfo{volume}{2}. IEEE, \bibinfo{pages}{1475--1480}.
\newblock


\bibitem[\protect\citeauthoryear{Sutton and Barto}{Sutton and Barto}{2018}]%
        {sutton2018reinforcement}
\bibfield{author}{\bibinfo{person}{Richard~S Sutton} {and}
  \bibinfo{person}{Andrew~G Barto}.} \bibinfo{year}{2018}\natexlab{}.
\newblock \bibinfo{booktitle}{\emph{Reinforcement learning: An introduction}}.
\newblock \bibinfo{publisher}{MIT press}.
\newblock


\bibitem[\protect\citeauthoryear{Thomas, Chien, Tamar, Ojea, and Abbeel}{Thomas
  et~al\mbox{.}}{2018}]%
        {thomas2018learning}
\bibfield{author}{\bibinfo{person}{Garrett Thomas}, \bibinfo{person}{Melissa
  Chien}, \bibinfo{person}{Aviv Tamar}, \bibinfo{person}{Juan~Aparicio Ojea},
  {and} \bibinfo{person}{Pieter Abbeel}.} \bibinfo{year}{2018}\natexlab{}.
\newblock \showarticletitle{Learning robotic assembly from cad}. In
  \bibinfo{booktitle}{\emph{2018 IEEE International Conference on Robotics and
  Automation (ICRA)}}. IEEE, \bibinfo{pages}{3524--3531}.
\newblock


\bibitem[\protect\citeauthoryear{Todorov, Erez, and Tassa}{Todorov
  et~al\mbox{.}}{2012}]%
        {todorov2012mujoco}
\bibfield{author}{\bibinfo{person}{Emanuel Todorov}, \bibinfo{person}{Tom
  Erez}, {and} \bibinfo{person}{Yuval Tassa}.} \bibinfo{year}{2012}\natexlab{}.
\newblock \showarticletitle{Mujoco: A physics engine for model-based control}.
  In \bibinfo{booktitle}{\emph{2012 IEEE/RSJ international conference on
  intelligent robots and systems}}. IEEE, \bibinfo{pages}{5026--5033}.
\newblock


\bibitem[\protect\citeauthoryear{Wan, Harada, and Nagata}{Wan
  et~al\mbox{.}}{2017}]%
        {wan2017assembly}
\bibfield{author}{\bibinfo{person}{Weiwei Wan}, \bibinfo{person}{Kensuke
  Harada}, {and} \bibinfo{person}{Kazuyuki Nagata}.}
  \bibinfo{year}{2017}\natexlab{}.
\newblock \showarticletitle{Assembly sequence planning for motion planning}.
\newblock \bibinfo{journal}{\emph{Assembly Automation}} (\bibinfo{year}{2017}).
\newblock


\bibitem[\protect\citeauthoryear{Wang, Rong, and Xiang}{Wang
  et~al\mbox{.}}{2014}]%
        {wang2014mechanical}
\bibfield{author}{\bibinfo{person}{Hui Wang}, \bibinfo{person}{Yiming Rong},
  {and} \bibinfo{person}{Dong Xiang}.} \bibinfo{year}{2014}\natexlab{}.
\newblock \showarticletitle{Mechanical assembly planning using ant colony
  optimization}.
\newblock \bibinfo{journal}{\emph{Computer-Aided Design}}  \bibinfo{volume}{47}
  (\bibinfo{year}{2014}), \bibinfo{pages}{59--71}.
\newblock


\bibitem[\protect\citeauthoryear{Wang, Zhou, Shi, Chen, Zhao, and Xu}{Wang
  et~al\mbox{.}}{2019b}]%
        {wang2019shape2motion}
\bibfield{author}{\bibinfo{person}{Xiaogang Wang}, \bibinfo{person}{Bin Zhou},
  \bibinfo{person}{Yahao Shi}, \bibinfo{person}{Xiaowu Chen},
  \bibinfo{person}{Qinping Zhao}, {and} \bibinfo{person}{Kai Xu}.}
  \bibinfo{year}{2019}\natexlab{b}.
\newblock \showarticletitle{Shape2motion: Joint analysis of motion parts and
  attributes from 3d shapes}. In \bibinfo{booktitle}{\emph{IEEE Conference on
  Computer Vision and Pattern Recognition (CVPR)}}.
  \bibinfo{pages}{8876--8884}.
\newblock


\bibitem[\protect\citeauthoryear{Wang, Weidner, Baxter, Hwang, Kaufman, and
  Sueda}{Wang et~al\mbox{.}}{2019a}]%
        {wang2019redmax}
\bibfield{author}{\bibinfo{person}{Ying Wang}, \bibinfo{person}{Nicholas~J
  Weidner}, \bibinfo{person}{Margaret~A Baxter}, \bibinfo{person}{Yura Hwang},
  \bibinfo{person}{Danny~M Kaufman}, {and} \bibinfo{person}{Shinjiro Sueda}.}
  \bibinfo{year}{2019}\natexlab{a}.
\newblock \showarticletitle{REDMAX: Efficient \& flexible approach for
  articulated dynamics}.
\newblock \bibinfo{journal}{\emph{ACM Transactions on Graphics (TOG)}}
  \bibinfo{volume}{38}, \bibinfo{number}{4} (\bibinfo{year}{2019}),
  \bibinfo{pages}{1--10}.
\newblock


\bibitem[\protect\citeauthoryear{Willis, Jayaraman, Chu, Tian, Li, Grandi,
  Sanghi, Tran, Lambourne, Solar-Lezama, and Matusik}{Willis
  et~al\mbox{.}}{2022}]%
        {willis2022joinable}
\bibfield{author}{\bibinfo{person}{Karl~DD Willis},
  \bibinfo{person}{Pradeep~Kumar Jayaraman}, \bibinfo{person}{Hang Chu},
  \bibinfo{person}{Yunsheng Tian}, \bibinfo{person}{Yifei Li},
  \bibinfo{person}{Daniele Grandi}, \bibinfo{person}{Aditya Sanghi},
  \bibinfo{person}{Linh Tran}, \bibinfo{person}{Joseph~G Lambourne},
  \bibinfo{person}{Armando Solar-Lezama}, {and} \bibinfo{person}{Wojciech
  Matusik}.} \bibinfo{year}{2022}\natexlab{}.
\newblock \showarticletitle{JoinABLe: Learning Bottom-up Assembly of Parametric
  CAD Joints}. In \bibinfo{booktitle}{\emph{Proceedings of the IEEE/CVF
  Conference on Computer Vision and Pattern Recognition (CVPR)}}.
\newblock


\bibitem[\protect\citeauthoryear{Willis, Pu, Luo, Chu, Du, Lambourne,
  Solar-Lezama, and Matusik}{Willis et~al\mbox{.}}{2021}]%
        {willis2021fusion}
\bibfield{author}{\bibinfo{person}{Karl D.~D. Willis}, \bibinfo{person}{Yewen
  Pu}, \bibinfo{person}{Jieliang Luo}, \bibinfo{person}{Hang Chu},
  \bibinfo{person}{Tao Du}, \bibinfo{person}{Joseph~G. Lambourne},
  \bibinfo{person}{Armando Solar-Lezama}, {and} \bibinfo{person}{Wojciech
  Matusik}.} \bibinfo{year}{2021}\natexlab{}.
\newblock \showarticletitle{Fusion 360 Gallery: A Dataset and Environment for
  Programmatic CAD Construction from Human Design Sequences}.
\newblock \bibinfo{journal}{\emph{ACM Transactions on Graphics (TOG)}}
  \bibinfo{volume}{40}, \bibinfo{number}{4} (\bibinfo{year}{2021}).
\newblock


\bibitem[\protect\citeauthoryear{Wilson and Latombe}{Wilson and
  Latombe}{1994}]%
        {wilson1994geometric}
\bibfield{author}{\bibinfo{person}{Randall~H Wilson} {and}
  \bibinfo{person}{Jean-Claude Latombe}.} \bibinfo{year}{1994}\natexlab{}.
\newblock \showarticletitle{Geometric reasoning about mechanical assembly}.
\newblock \bibinfo{journal}{\emph{Artificial Intelligence}}
  \bibinfo{volume}{71}, \bibinfo{number}{2} (\bibinfo{year}{1994}),
  \bibinfo{pages}{371--396}.
\newblock


\bibitem[\protect\citeauthoryear{Xiang, Qin, Mo, Xia, Zhu, Liu, Liu, Jiang,
  Yuan, Wang, Yi, Chang, Guibas, and Su}{Xiang et~al\mbox{.}}{2020}]%
        {xiang2020sapien}
\bibfield{author}{\bibinfo{person}{Fanbo Xiang}, \bibinfo{person}{Yuzhe Qin},
  \bibinfo{person}{Kaichun Mo}, \bibinfo{person}{Yikuan Xia},
  \bibinfo{person}{Hao Zhu}, \bibinfo{person}{Fangchen Liu},
  \bibinfo{person}{Minghua Liu}, \bibinfo{person}{Hanxiao Jiang},
  \bibinfo{person}{Yifu Yuan}, \bibinfo{person}{He Wang}, \bibinfo{person}{Li
  Yi}, \bibinfo{person}{Angel~X. Chang}, \bibinfo{person}{Leonidas~J. Guibas},
  {and} \bibinfo{person}{Hao Su}.} \bibinfo{year}{2020}\natexlab{}.
\newblock \showarticletitle{{SAPIEN}: A SimulAted Part-based Interactive
  ENvironment}. In \bibinfo{booktitle}{\emph{IEEE Conference on Computer Vision
  and Pattern Recognition (CVPR)}}.
\newblock


\bibitem[\protect\citeauthoryear{Xu, Chen, Zlokapa, Foshey, Matusik, Sueda, and
  Agrawal}{Xu et~al\mbox{.}}{2021}]%
        {xu2021end}
\bibfield{author}{\bibinfo{person}{Jie Xu}, \bibinfo{person}{Tao Chen},
  \bibinfo{person}{Lara Zlokapa}, \bibinfo{person}{Michael Foshey},
  \bibinfo{person}{Wojciech Matusik}, \bibinfo{person}{Shinjiro Sueda}, {and}
  \bibinfo{person}{Pulkit Agrawal}.} \bibinfo{year}{2021}\natexlab{}.
\newblock \showarticletitle{{An End-to-End Differentiable Framework for
  Contact-Aware Robot Design}}. In \bibinfo{booktitle}{\emph{Proceedings of
  Robotics: Science and Systems}}. \bibinfo{address}{Virtual}.
\newblock
\urldef\tempurl%
\url{https://doi.org/10.15607/RSS.2021.XVII.008}
\showDOI{\tempurl}


\bibitem[\protect\citeauthoryear{Yan, Hu, Yan, Chen, van Kaick, Zhang, and
  Huang}{Yan et~al\mbox{.}}{2019}]%
        {yanRPMNet19}
\bibfield{author}{\bibinfo{person}{Zihao Yan}, \bibinfo{person}{Ruizhen Hu},
  \bibinfo{person}{Xingguang Yan}, \bibinfo{person}{Luanmin Chen},
  \bibinfo{person}{Oliver van Kaick}, \bibinfo{person}{Hao Zhang}, {and}
  \bibinfo{person}{Hui Huang}.} \bibinfo{year}{2019}\natexlab{}.
\newblock \showarticletitle{RPM-Net: Recurrent Prediction of Motion and Parts
  from Point Cloud}.
\newblock \bibinfo{journal}{\emph{Annual Conference on Computer Graphics and
  Interactive Techniques Asia (SIGGRAPH Asia)}} \bibinfo{volume}{38},
  \bibinfo{number}{6} (\bibinfo{year}{2019}), \bibinfo{pages}{240:1--240:15}.
\newblock


\bibitem[\protect\citeauthoryear{Yu, Shao, Chen, Wu, Fan, Mo, and Dong}{Yu
  et~al\mbox{.}}{2021}]%
        {yu2021roboassembly}
\bibfield{author}{\bibinfo{person}{Mingxin Yu}, \bibinfo{person}{Lin Shao},
  \bibinfo{person}{Zhehuan Chen}, \bibinfo{person}{Tianhao Wu},
  \bibinfo{person}{Qingnan Fan}, \bibinfo{person}{Kaichun Mo}, {and}
  \bibinfo{person}{Hao Dong}.} \bibinfo{year}{2021}\natexlab{}.
\newblock \showarticletitle{RoboAssembly: Learning Generalizable Furniture
  Assembly Policy in a Novel Multi-robot Contact-rich Simulation Environment}.
\newblock \bibinfo{journal}{\emph{arXiv preprint arXiv:2112.10143}}
  (\bibinfo{year}{2021}).
\newblock


\bibitem[\protect\citeauthoryear{Zakka, Zeng, Lee, and Song}{Zakka
  et~al\mbox{.}}{2020}]%
        {zakka2020form2fit}
\bibfield{author}{\bibinfo{person}{Kevin Zakka}, \bibinfo{person}{Andy Zeng},
  \bibinfo{person}{Johnny Lee}, {and} \bibinfo{person}{Shuran Song}.}
  \bibinfo{year}{2020}\natexlab{}.
\newblock \showarticletitle{Form2fit: Learning shape priors for generalizable
  assembly from disassembly}. In \bibinfo{booktitle}{\emph{2020 IEEE
  International Conference on Robotics and Automation (ICRA)}}. IEEE,
  \bibinfo{pages}{9404--9410}.
\newblock


\bibitem[\protect\citeauthoryear{Zhang, Belfer, Kry, and Vouga}{Zhang
  et~al\mbox{.}}{2020}]%
        {zhang2020c}
\bibfield{author}{\bibinfo{person}{Xinya Zhang}, \bibinfo{person}{Robert
  Belfer}, \bibinfo{person}{Paul~G Kry}, {and} \bibinfo{person}{Etienne
  Vouga}.} \bibinfo{year}{2020}\natexlab{}.
\newblock \showarticletitle{C-Space tunnel discovery for puzzle path planning}.
\newblock \bibinfo{journal}{\emph{ACM Transactions on Graphics (TOG)}}
  \bibinfo{volume}{39}, \bibinfo{number}{4} (\bibinfo{year}{2020}),
  \bibinfo{pages}{104--1}.
\newblock


\bibitem[\protect\citeauthoryear{Zhao}{Zhao}{2005}]%
        {zhao2005fast}
\bibfield{author}{\bibinfo{person}{Hongkai Zhao}.}
  \bibinfo{year}{2005}\natexlab{}.
\newblock \showarticletitle{A fast sweeping method for eikonal equations}.
\newblock \bibinfo{journal}{\emph{Mathematics of computation}}
  \bibinfo{volume}{74}, \bibinfo{number}{250} (\bibinfo{year}{2005}),
  \bibinfo{pages}{603--627}.
\newblock


\bibitem[\protect\citeauthoryear{Zhou, Sueda, Matusik, and Shamir}{Zhou
  et~al\mbox{.}}{2014}]%
        {Zhou2014}
\bibfield{author}{\bibinfo{person}{Yahan Zhou}, \bibinfo{person}{Shinjiro
  Sueda}, \bibinfo{person}{Wojciech Matusik}, {and} \bibinfo{person}{Ariel
  Shamir}.} \bibinfo{year}{2014}\natexlab{}.
\newblock \showarticletitle{Boxelization: Folding 3D Objects into Boxes}.
\newblock \bibinfo{journal}{\emph{ACM Trans. Graph.}} \bibinfo{volume}{33},
  \bibinfo{number}{4}, Article \bibinfo{articleno}{71} (\bibinfo{date}{Jul}
  \bibinfo{year}{2014}), \bibinfo{numpages}{8}~pages.
\newblock
\showISSN{0730-0301}


\bibitem[\protect\citeauthoryear{Zhu and Hu}{Zhu and Hu}{2018}]%
        {zhu2018robot}
\bibfield{author}{\bibinfo{person}{Zuyuan Zhu} {and} \bibinfo{person}{Huosheng
  Hu}.} \bibinfo{year}{2018}\natexlab{}.
\newblock \showarticletitle{Robot learning from demonstration in robotic
  assembly: A survey}.
\newblock \bibinfo{journal}{\emph{Robotics}} \bibinfo{volume}{7},
  \bibinfo{number}{2} (\bibinfo{year}{2018}), \bibinfo{pages}{17}.
\newblock


\bibitem[\protect\citeauthoryear{Zickler and Veloso}{Zickler and
  Veloso}{2009}]%
        {zickler2009efficient}
\bibfield{author}{\bibinfo{person}{Stefan Zickler} {and}
  \bibinfo{person}{Manuela~M Veloso}.} \bibinfo{year}{2009}\natexlab{}.
\newblock \showarticletitle{Efficient physics-based planning: sampling search
  via non-deterministic tactics and skills.}. In
  \bibinfo{booktitle}{\emph{AAMAS (1)}}. Citeseer, \bibinfo{pages}{27--33}.
\newblock


\end{thebibliography}

\newpage
\begin{appendices}
    \section{SDF Construction in Simulation}
\label{supp:sdf}

We construct the SDF grid for an object using the Fast Sweeping Algorithm \cite{zhao2005fast} once at the initialization stage of the simulation. The Fast Sweeping Algorithm computes the distance function at the grid points around the object surface, and then updates the remaining grids with eight sweeps. Given the SDF grid, the distance of an arbitrary point in the space can acquired in $\mathcal{O}(1)$ by trilinearly interpolating among surrounding grid cells and the gradients are computed in $\mathcal{O}(1)$ via finite-differencing. Our assembly dataset contains thousands of objects with dramatically different sizes and also contains objects having different sizes in different dimensions (e.g. long poles). In order to capture the surface texture details of those objects while keeping the algorithm as fast and low-memory as possible, we use an adaptive SDF grid with cell size $\min(0.05, L_i/20)$ in each dimension $i$ for an object, where $L_i$ is the length of the object along the $i$-th dimension and assuming each assembly is scaled to fit in a 10x10x10 unit cube.

\section{Dataset Details}

\subsection{Dataset Pre-Processing}
\label{supp:data_process}

Our dataset pre-processing consists of several steps to adapt the previous CAD assembly datasets for assembly planning. Specifically, in each assembly that contains two or multiple parts (meshes), we sequentially do the following steps:

\begin{enumerate}
    \item Load meshes from the source datasets with correct transforms to the assembled states applied.
    \item Remove non-watertight meshes since the distance query is poorly defined on non-watertight meshes.
    \item Remove duplicate meshes with the same geometry and position as other meshes, which is a rare design mistake.
    \item Remove meshes that overlap more than 10\% with other meshes because assembly planning will less likely give a plausible result for such a large penetration. The overlap percentage is computed by sampling points inside the mesh volume and checking the ratio of points contained by other meshes.
    \item Remove thin meshes, which are difficult for SDF-based collision detection to compute distance correctly. This step is optional if the simulation is not based on SDF for collision detection and can handle thin structures well. In our implementation, we first compute the mesh's oriented bounding box (OBB). Next, we filter out meshes whose thinnest edge of OBB is smaller than 1\% of the largest \textit{scale} of the whole assembly, where the \textit{scale} is defined by the length of the largest edge of the assembly's bounding box.
    \item Remove all other parts except the largest \textit{connected} subset of the assembly. For example, all single floating parts will be removed. This step makes the assembly data more meaningful for evaluating assembly planning methods because parts floating in the air are already disassembled or very close to being disassembled. In our implementation, two \textit{connected} parts are defined as having a collision between their convex hulls.
    \item Normalize all the remaining meshes in an assembly to fit in a 10x10x10 bounding box. This makes all assemblies have similar scales, thus benchmarking assembly planning methods on the whole dataset becomes much easier, considering setting a single set of hyper-parameters of the algorithm that generalize to the entire dataset (e.g., collision detection threshold, state similarity threshold, step size of RRT).
    \item Remove assemblies that are not disassemblable as rigid parts. This is done manually since an oracle does not exist to check if an assembly is disassemblable. This step is not guaranteed to be highly accurate considering the manual nature and dataset size.
    \item Subdivide the meshes such that the largest edge of the mesh is no larger than $0.5$. This is because our simulation relies on a point-based contact model and does not support edge-edge collision detection yet. Increasing the density of mesh vertices helps the simulation handle contacts more robustly. This step is optional if the simulation can handle edge-edge collision well.
\end{enumerate}

Note that we do not remove meshes with slight overlaps (penetration) with other meshes. This is because assembly CAD models designed by human designers typically have small penetration errors between parts. Rather than requiring the assembly model to be perfectly penetration-free, we set an adaptive threshold for collision detection based on the amount of initial penetration (for both our method and baseline methods). This allows path planners to generate plausible disassembly paths while some slight penetration exists.

\subsection{Rotational Dataset Overview}
\label{supp:rot_data}

\begin{figure*}
    \centering
    \includegraphics[width=0.7\textwidth]{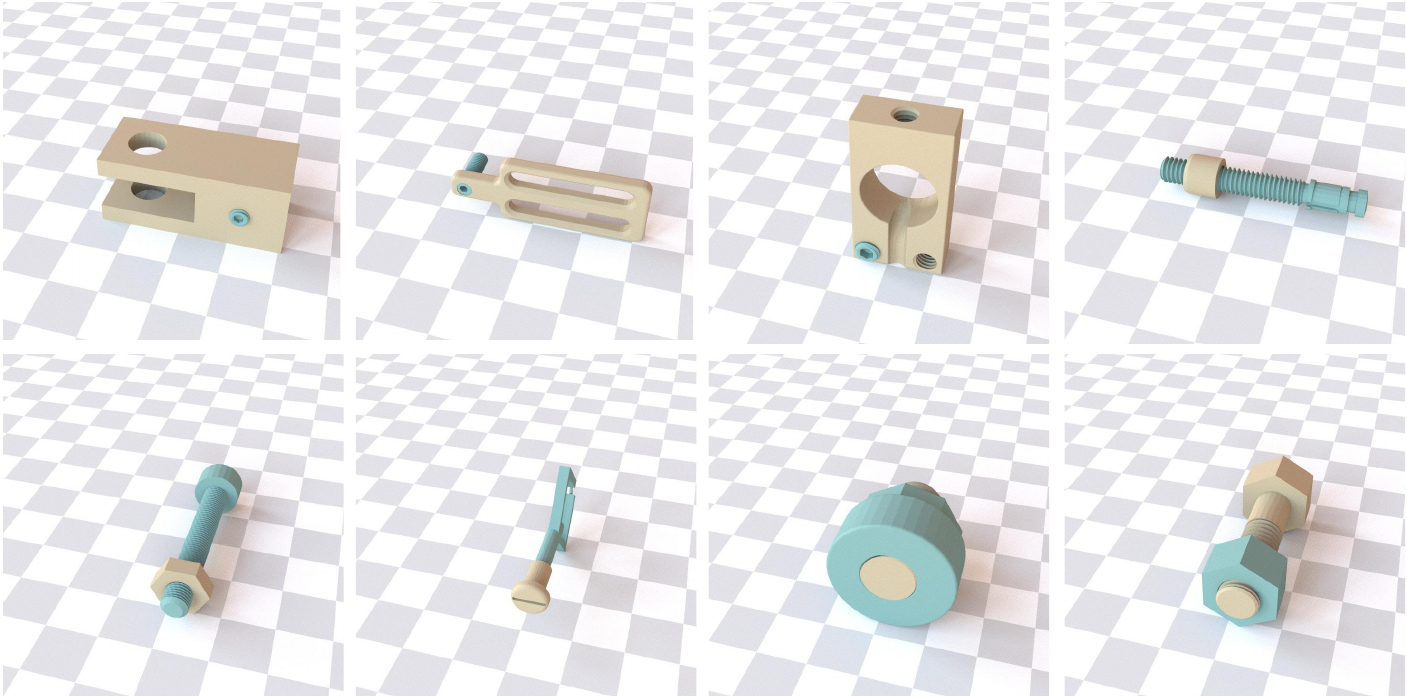}
    \vspace{-2mm}
    \caption{\textbf{Screw} category of the rotational dataset.}
    \label{fig:rotation_dataset_screw}
\end{figure*}

\begin{figure*}
    \centering
    \includegraphics[width=0.7\textwidth]{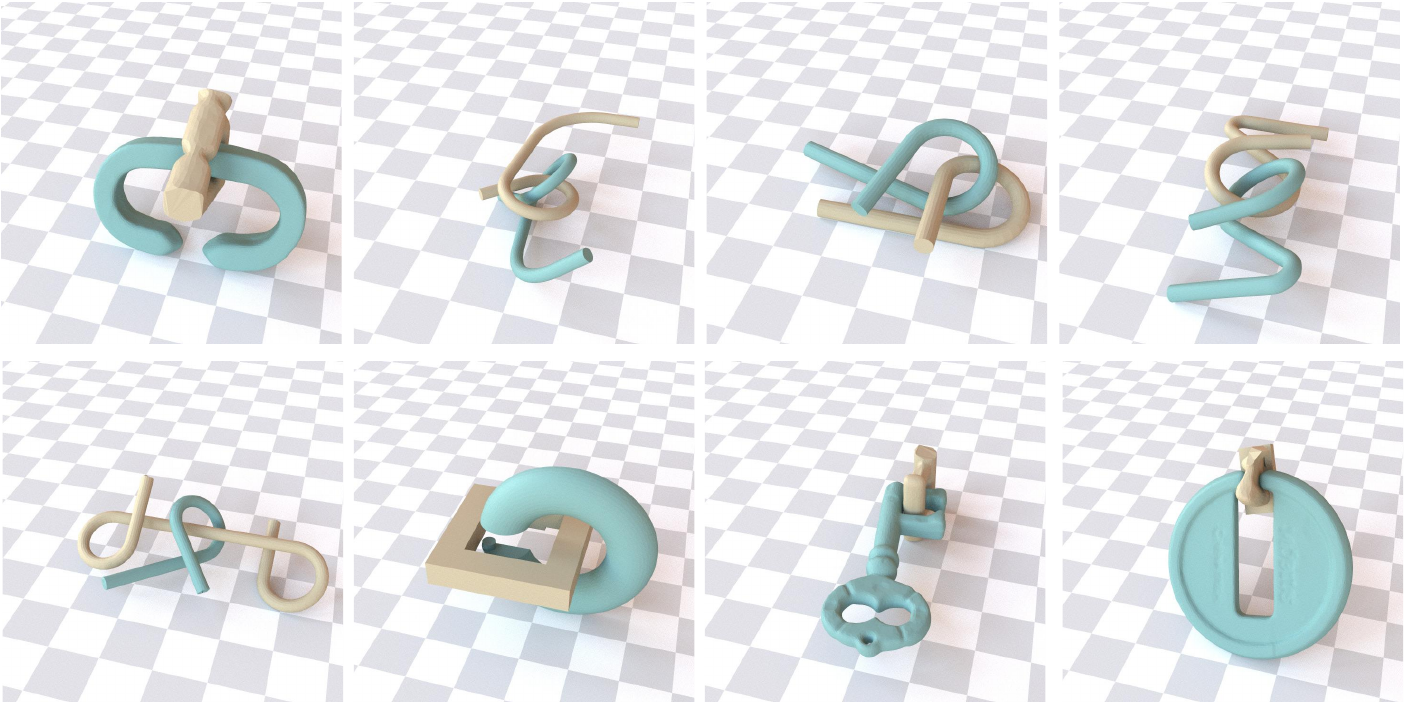}
    \vspace{-2mm}
    \caption{\textbf{Puzzle} category of the rotational dataset.}
    \label{fig:rotation_dataset_puzzle}
\end{figure*}

\begin{figure*}
    \centering
    \includegraphics[width=0.7\textwidth]{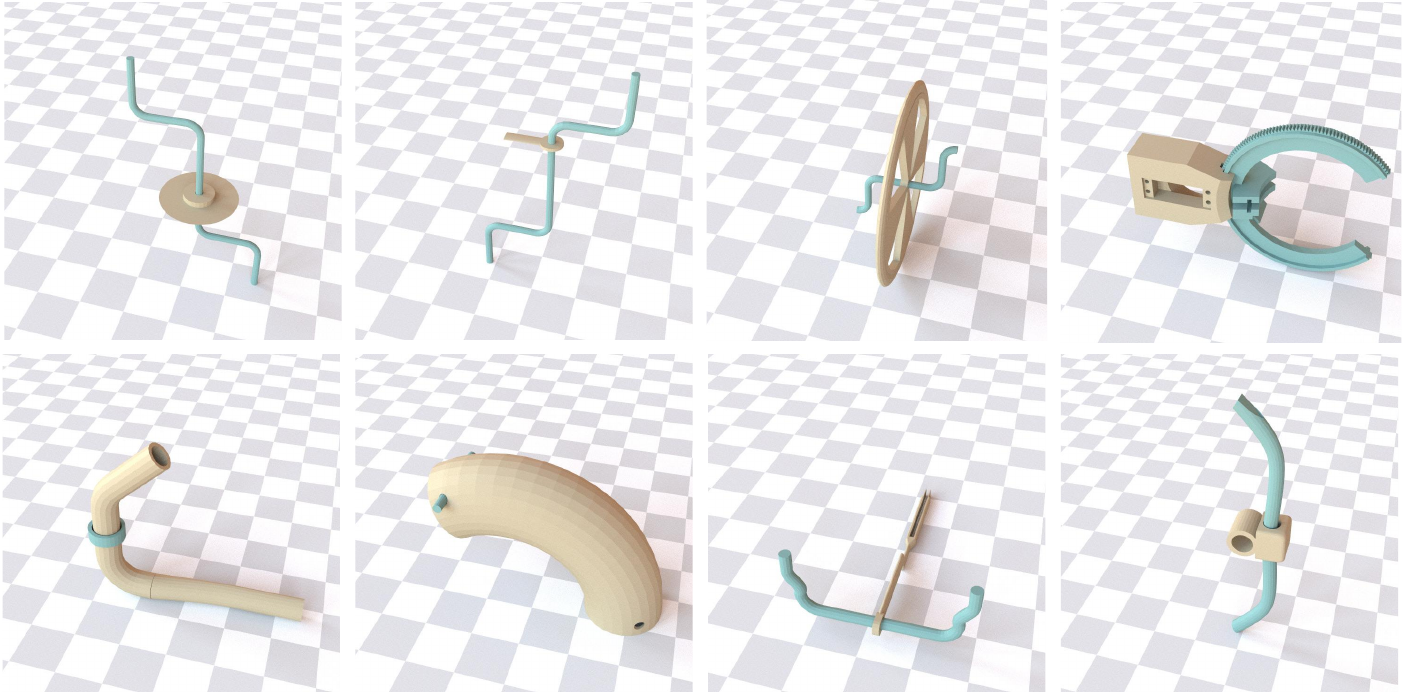}
    \vspace{-2mm}
    \caption{\textbf{Others} category of the rotational dataset.}
    \label{fig:rotation_dataset_others}
\end{figure*}

Figure~\ref{fig:rotation_dataset_screw}, \ref{fig:rotation_dataset_puzzle} and \ref{fig:rotation_dataset_others} show a complete overview of our rotational dataset, which consists of rotational assemblies that require various types of assembly motion.

\section{Experimental Setup}

\subsection{Baseline Methods}
\label{supp:algo}

In this section, we describe some of our baseline methods in more details:

\textbf{RRT}~\cite{lavalle1998rapidly}: RRT iteratively grows a tree in the state space to reach the user-defined goal state. In Each iteration it either samples a random state in the space or chooses the goal state and then extends towards it from the nearest node in the existing tree. However, it requires an explicitly specified goal state, which can be an arbitrary disassembled state in disassembly path planning. Therefore, we use a randomly selected disassembled state as the goal for RRT.

\textbf{T-RRT}~\cite{aguinaga2008parallel}: Instead of choosing a random disassembled goal state like RRT, when expanding the tree, Targetless-RRT takes as the goal the nearest state outside the bounding box of the rest of the assembly. This goal heuristic is empirically more suitable for disassembly than random.

\textbf{BK-RRT}~\cite{zickler2009efficient}: See Algorithm~\ref{alg:bk-rrt} for a complete illustration of BK-RRT in disassembly path planning.

\begin{algorithm}[h]
\SetAlgoLined
\KwIn{Assembled state $s_0$, timeout $t_{\max}$, simulation time step $\Delta t$.}
\KwOut{A disassembly path $P_D = \{s_0,...,s_n\}$.}
$T$ = EmptyTree()\;
$T$.AddNode($s_0$)\;
\While{$t$ < $t_{\max}$}{
    $s_r$ = SampleRandomState()\;
    $s_i$ = NearestNeighbor($T$, $s_r$)\;
    $a_i$ = RandomAction()\;
    $s_{i+1}$ = Simulate($s_i$, $a_i$, $\Delta t$)\;
    $T$.AddNode($s_{i+1}$)\;
    $T$.AddEdge($s_i$, $s_{i+1}$, $a_i$)\;
    \If{IsDisassembled($s_{i+1}$)}{
        \Return{$T$.GetPath($s_0$, $s_{i+1}$)}\;
    }
}
\Return{failed}\;
\caption{BK-RRT for Disassembly Path Planning}
\label{alg:bk-rrt}
\end{algorithm}

\subsection{Hyper-Parameters}
\label{supp:hp}

In this section, we present all hyper-parameters used in experimental evaluation. Specifically, Table~\ref{tab:hp_sim} summarizes the main hyper-parameters used in our physics-based simulation, Table~\ref{tab:hp_physics} shows the hyper-parameters for physics-based path planners (our method and BK-RRT), and Table~\ref{tab:hp_geometric} shows the hyper-parameters for geometric-based path planners (RRT, T-RRT, MV+T-RRT).

\begin{table}%
\small
\caption{Hyper-parameters of physics-based simulation.}
\label{tab:hp_sim}
\begin{center}
\begin{tabular}{l|c}
  \toprule
  \textbf{Name} & \textbf{Value} \\
  \midrule
  Contact stiffness $k_n$ & 1e6 \\
  Contact damping coefficient $k_d$ & 0 \\
  Simulation time step & 1e-3 \\
  \bottomrule

\end{tabular}
\end{center}
\end{table}%

\begin{table}%
\small
\caption{Hyper-parameters of physics-based path planners.}
\label{tab:hp_physics}
\begin{center}
\begin{tabular}{l|c}
  \toprule
  \textbf{Name} & \textbf{Value} \\
  \midrule
  Path planning time step $\Delta_t$ & 1e-1 \\
  Penetration threshold for collision detection & 0.01 \\
  Force/torque magnitude of each action & 100 \\
  State similarity threshold (translation) $\delta_t$ & 0.05 \\
  State similarity threshold (rotation) $\delta_r$ & 0.5 \\
  \bottomrule

\end{tabular}
\end{center}
\end{table}%

\begin{table}%
\small
\caption{Hyper-parameters of geometric-based path planners.}
\label{tab:hp_geometric}
\begin{center}
\begin{tabular}{l|c}
  \toprule
  \textbf{Name} & \textbf{Value} \\
  \midrule
  Step size of tree extension & 0.01 \\
  Maximum penetration allowed & 0.01 \\
  Goal probability of T-RRT & 0.2 \\
  
  \bottomrule

\end{tabular}
\end{center}
\end{table}%

\section{More Results}
\label{supp:results}

\begin{figure*}
    \centering
    \includegraphics[width=0.9\textwidth]{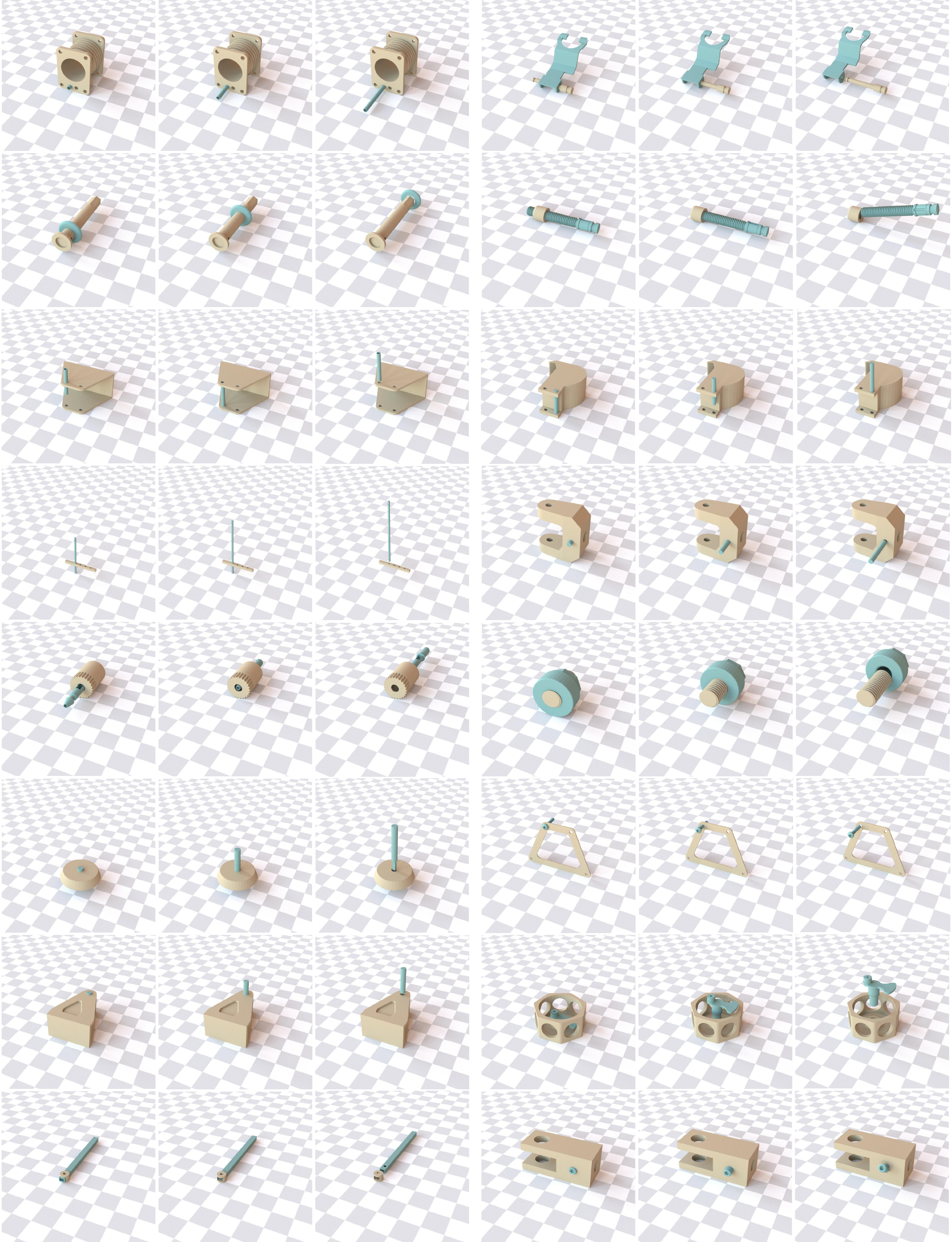}
    \caption{More successful examples of two-part disassembly motion produced by our method while failing on baseline methods.}
    \label{fig:twopart_motion_appendix}
\end{figure*}

See Figure~\ref{fig:twopart_motion_appendix} for more assembly motion produced by our method which have a low success rate on other baseline methods.

\end{appendices}

\end{document}